\PassOptionsToPackage{table}{xcolor}
\documentclass{fairmeta}

\usepackage{amsmath,amssymb}
\usepackage{adjustbox}
\usepackage{float}
\usepackage{pifont}
\usepackage{tabularx}
\usepackage{titletoc}

\definecolor{resultAzure}{HTML}{CFE3F1}
\definecolor{deltaApricot}{HTML}{F8E9DC}
\definecolor{sectionMist}{HTML}{EBF2F7}
\definecolor{ourblue}{HTML}{DDF0EB}
\definecolor{recasttablehighlight}{HTML}{DDF0EB}
\newcommand{\abldrop}[2]{%
  \mbox{#1\textsuperscript{\textcolor{gray}{\scriptsize\ensuremath{(-#2)}}}}%
}

\ifdefined\pdfobjcompresslevel
\fi

\title{RECAST: Training-Based Recursive Self-Improvement in VideoLLM}
\renewcommand\authorformat[2][]{\mbox{\sffamily\bfseries #2$^{#1}$}}
\renewcommand\affiliationformat[2][]{\mbox{\small $^{#1}$#2}}
\newcommand\authornewrow[2][]{\addtolist[#1]{#2}{\authorlist}{\authorformat}{,\protect\newline}}
\newcommand\affiliationnewrow[2][]{\addtolist[#1]{#2}{\affiliationlist}{\affiliationformat}{,\protect\newline}}

\author[1,2,\dagger]{Weijie Zhu}
\author[3,\dagger]{Yuzhe Zhang}
\author[1,\dagger]{Siyi Guo}
\author[3]{Anzhi L{\"u}}
\author[1]{Hanyu Fu}
\author[1]{Tianyi Ding}
\author[1]{Feiran Liu}
\authornewrow[4]{Rongpeng Zhu}
\author[1]{Xin Wei}
\author[1]{Zhiyu Lin}
\author[1]{Hongbo Sun}
\author[1]{Ye Yuan}
\author[1,*]{Han Fang}
\author[1,*]{Hao Sun}

\affiliation[1]{China Telecom Artificial Intelligence Technology (Beijing) Co., Ltd.}
\affiliationnewrow[2]{University of Chinese Academy of Sciences}
\affiliation[3]{Peking University}
\affiliation[4]{Shanghai Jiao Tong University}
\contribution[\dagger]{Equal contribution}
\contribution[*]{Corresponding authors.}

\metadata[Project page]{\href{https://whiteventral.github.io/RECAST/}{%
  \raisebox{-2pt}{\includegraphics[height=10pt]{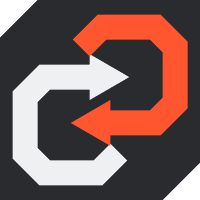}}\hspace{0.45em}\nolinkurl{https://whiteventral.github.io/RECAST/}}}
\patchcmd{\mymaketitle}{\vskip 0.5cm}{\vskip 0.12cm}{}{%
  \ClassError{fairmeta}{Cannot set project-link spacing}{Check the title-box definition.}}
\patchcmd{\mymaketitle}{\setlength{\parskip}{0.5cm}}{\setlength{\parskip}{0.2cm}}{}{%
  \ClassError{fairmeta}{Cannot set title-box paragraph spacing}{Check the title-box definition.}}
\patchcmd{\mymaketitle}{\tcbset{bottom=0.35cm}}{\tcbset{bottom=0.15cm}}{}{%
  \ClassError{fairmeta}{Cannot set title-box bottom padding}{Check the title-box definition.}}
\hypersetup{
  pdftitle={RECAST: Training-Based Recursive Self-Improvement in VideoLLM},
  pdfauthor={Weijie Zhu, Yuzhe Zhang, Siyi Guo, Anzhi L{\"u}, Hanyu Fu, Tianyi Ding, Feiran Liu, Rongpeng Zhu, Xin Wei, Zhiyu Lin, Hongbo Sun, Ye Yuan, Han Fang, Hao Sun}
}

\begin{document}
\begingroup
\setlength{\emergencystretch}{1em}
\setlength{\parskip}{4pt plus 1pt}
\setlength{\textfloatsep}{12pt plus 2pt minus 2pt}
\setlength{\floatsep}{10pt plus 2pt minus 2pt}
\setlength{\intextsep}{10pt plus 2pt minus 2pt}
\setlength{\abovedisplayskip}{7pt plus 2pt minus 2pt}
\setlength{\belowdisplayskip}{7pt plus 2pt minus 2pt}
\setlength{\abovedisplayshortskip}{3pt plus 1pt}
\setlength{\belowdisplayshortskip}{5pt plus 1pt}
\titlespacing*{\section}{0pt}{12pt plus 2pt minus 2pt}{6pt plus 1pt}
\titlespacing*{\subsection}{0pt}{9pt plus 2pt minus 2pt}{4pt plus 1pt}
\captionsetup{skip=4pt}
\abstract{
Reliable video reasoning requires fine-grained feedback beyond sparse task rewards and local teacher preferences. We propose \textbf{RECAST} (\textbf{R}ecursive \textbf{E}volution through \textbf{C}hunk-\textbf{A}ware \textbf{S}tudent \textbf{T}raining), a training-based recursive self-improvement framework alternating reward learning and teacher-guided optimization to renew an evidence-conditioned self-teacher from reward-induced policy changes. \emph{Future-Aware Semantic Credit} (FASC) reweights sampled-token advantages using feedback from current and later semantic chunks, while \emph{Outcome-Grounded Decision Refinement} (ODR) refines alternatives through reward-scored student continuations. We construct \textbf{RECAST-100k}, a 100,000-example dataset enriched with spatiotemporal evidence to support this training. Across nine benchmarks in video reasoning, temporal grounding, and general video understanding, \textbf{RECAST-7B} leads all compared open-source models, averaging \textbf{50.8} in reasoning-and-grounding and \textbf{65.0} in general-video understanding; ablations confirm complementary benefits and sustained teacher-renewal gains, while improvements across tasks and model scales highlight RECAST as a broadly applicable training path for recursive self-improvement in VideoLLMs.
}

\maketitle

\begingroup
\setlength{\intextsep}{6pt}
\enlargethispage{\baselineskip}
\begin{figure}[H]
    \centering

    \includegraphics[
        width=0.95\linewidth
    ]{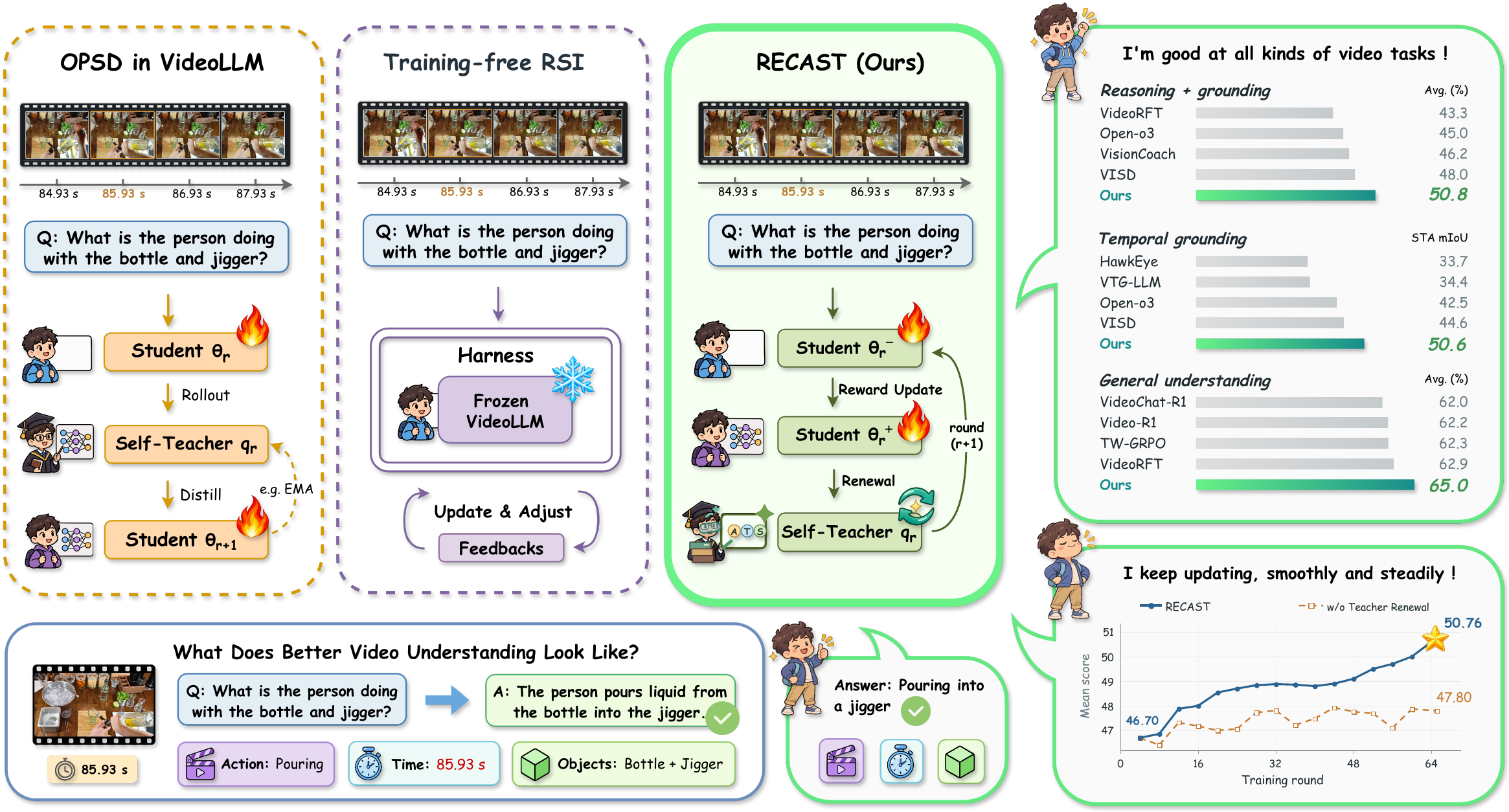}

    \caption{
        \textbf{RECAST for video understanding.}
        \textbf{Left:}
        Unlike OPSD and training-free RSI,
        RECAST alternates reward-driven student
        updates and self-teacher renewal to produce
        spatiotemporally grounded responses.
        \textbf{Right:}
        Results across three video-task groups and
        training-round comparisons demonstrate
        strong performance and the benefit of
        self-teacher renewal.
    }

    \label{fig:video-rsi-motivation}

\end{figure}
\endgroup

\section{Introduction}
\label{sec:intro}

Recent advances in Video Large Language Models (VideoLLMs) have improved video understanding and spatiotemporal reasoning~\citep{bai2025qwen,li2025videochatr1,meng2026openo3}. Yet, reasoning over complex video events requires both fine-grained guidance for intermediate decisions and task-level feedback for continuous improvement. Reinforcement learning with verifiable rewards (RLVR) provides outcome-level supervision~\citep{feng2025videor1,wang2025timer1,wang2025videorft}, while on-policy distillation (OPD) and on-policy self-distillation (OPSD) introduce teacher preferences for finer-grained optimization~\citep{li2026videoopd,wang2026videoopsd,zhu2026mesd,lin2026visd}. Meanwhile, training-free recursive self-improvement (RSI) enhances external reasoning harnesses without updating VideoLLM parameters~\citep{xu2026videoharness,luo2026videorsi,lin2026videoharnesssrsi}. As illustrated in Fig.~\ref{fig:video-rsi-motivation} (left), these paradigms do not directly integrate reward-driven teacher renewal into recursive parameter optimization. Although training-based RSI has been explored for LLMs~\citep{li2026rise}, extending it to VideoLLMs presents two key challenges.

\noindent\textbf{Challenge 1: How can preference credit respect semantic dependencies?}
Video reasoning involves interdependent objects, actions, timestamps, and spatial relations~\citep{meng2026openo3}. Local token-level preferences may overlook dependencies spanning multiple response segments. Effective guidance should therefore group related content into semantic chunks and account for subsequent reasoning when assigning credit to earlier decisions.

\noindent\textbf{Challenge 2: How can task outcomes refine teacher preferences?}
Teacher--student disagreement identifies alternative decisions but does not reveal which ones improve task performance. Moreover, rewards on sampled trajectories cannot evaluate unchosen alternatives. Reliable preference refinement thus requires testing candidate decisions through reward-evaluated continuations.

To address these challenges, we propose \textbf{RECAST} (\textbf{R}ecursive \textbf{E}volution through \textbf{C}hunk-\textbf{A}ware \textbf{S}tudent \textbf{T}raining), a training-based RSI framework that alternates reward learning and teacher-guided optimization, recursively renewing an evidence-conditioned self-teacher through reward-induced policy changes. We introduce two complementary mechanisms: \emph{Future-Aware Semantic Credit}, which incorporates feedback from current and subsequent semantic chunks to reweight sampled-token advantages; and \emph{Outcome-Grounded Decision Refinement}, which evaluates alternatives at teacher--student disagreement positions using reward-scored student continuations. Together, they combine semantic preference guidance with outcome-grounded refinement to support recursive student improvement.

We further construct \emph{RECAST-100k}, covering general video QA, spatiotemporally grounded QA, and temporal localization, with annotations supporting reward evaluation and evidence-conditioned guidance. As shown in Fig.~\ref{fig:video-rsi-motivation} (right), RECAST achieves strong performance across diverse video tasks, while training-round comparisons demonstrate the benefit of self-teacher renewal.

\begin{figure}[t]
    \centering
    \includegraphics[width=0.9\textwidth]{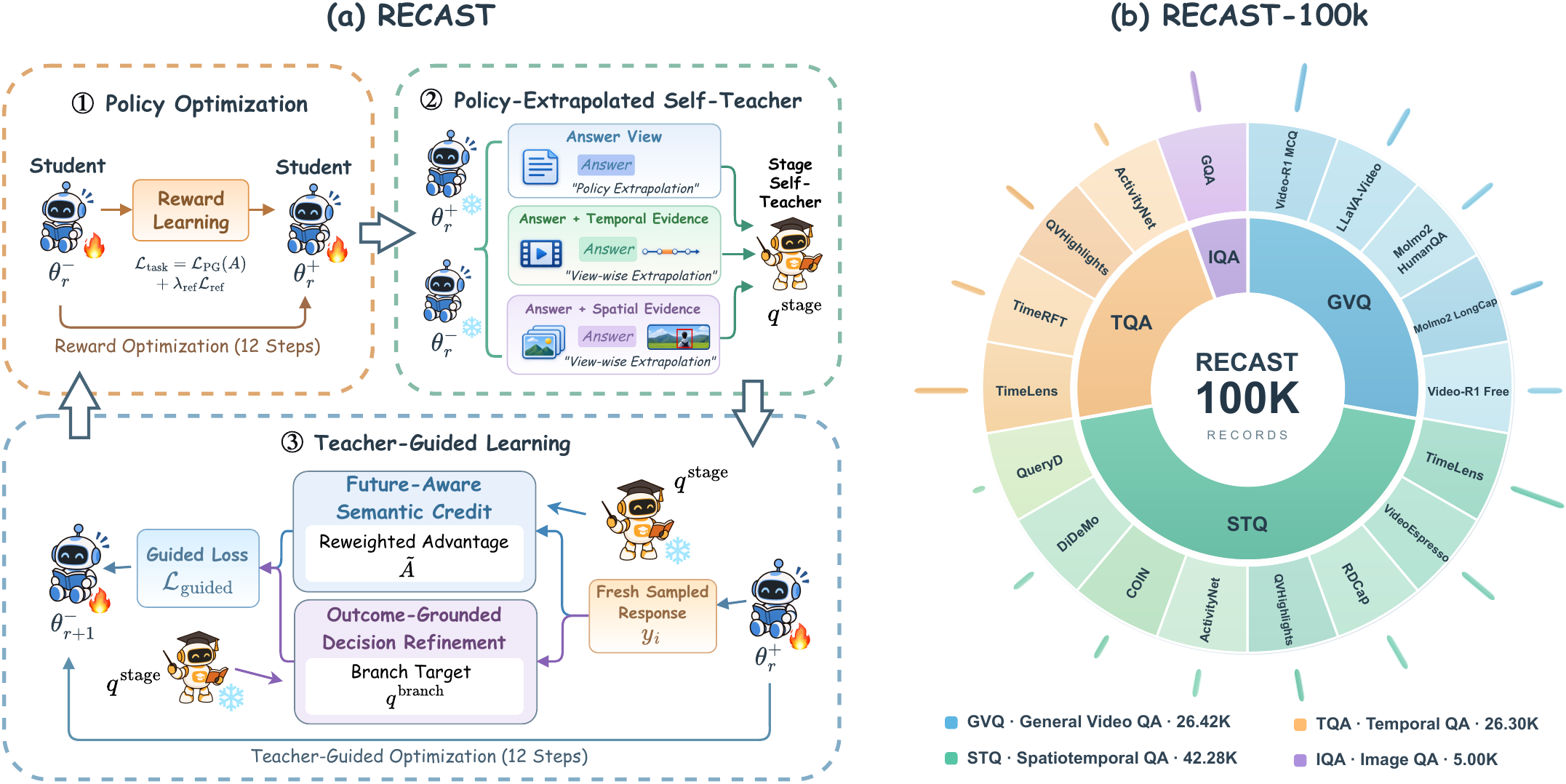}
    \caption{\textbf{RECAST and RECAST-100k overview.} Left: recursive student--teacher training. Right: task and source composition.}
    \label{fig:video-rsi-overview}
\end{figure}

Our main contributions are:
\begin{itemize}
    \item We propose \emph{RECAST}, a training-based RSI paradigm for VideoLLMs that renews evidence-conditioned teacher preferences across rounds and combines coherent semantic guidance with task feedback on alternative choices, enabling VideoLLMs to improve through iterative training.
    \item We construct \emph{RECAST-100k}, a dataset enriched with fine-grained spatiotemporal evidence to support alternating reward learning and self-distillation in training-based RSI for VideoLLMs.
    \item Across nine benchmarks, \emph{RECAST-7B} leads all compared open-source models in video reasoning, temporal grounding, and general video understanding, achieving a 50.8 reasoning-and-grounding average, leading all Charades-STA metrics, and reaching a 65.0 general-video average.
\end{itemize}
\section{Related Work}
\label{sec:related}

\noindent\textbf{Post-Training for VideoLLMs.}
Post-training for Video Large Language Models (VideoLLMs) has progressed from outcome-based policy optimization toward finer-grained supervision grounded in spatiotemporal evidence. Video-R1~\citep{feng2025videor1} and VideoRFT~\citep{wang2025videorft} improve video reasoning with reinforcement learning, while Open-o3 Video~\citep{meng2026openo3} combines answer correctness and spatiotemporal evidence in verifiable rewards. Yet such training remains largely driven by response-level rewards, which do not distinguish the contributions of intermediate judgments. To provide finer-grained signals, Video-OPD~\citep{li2026videoopd} applies on-policy distillation to temporal localization, using teacher distributions to guide student-generated trajectories; Video-OPSD~\citep{wang2026videoopsd} constructs a self-teacher with task-relevant visual frames. MeSD~\citep{zhu2026mesd} further uses teacher feedback conditioned on answer, temporal, and spatial evidence, while VISD~\citep{lin2026visd} introduces structured diagnostic information for local supervision in video reasoning. These studies make teacher feedback more fine-grained and more closely connected to visual evidence, but how to allocate that feedback across interrelated generated content remains an open issue. To address this, RECAST introduces \emph{Future-Aware Semantic Credit}: it groups related event information into semantic chunks and incorporates feedback from subsequent chunks when weighting sampled tokens, accounting for both within-event semantics and later reasoning.

\noindent\textbf{Recursive Self-Improvement for Video Reasoning.}
Recursive self-improvement (RSI) reuses a model's or system's own experience to guide further learning. For general language models, SPIN~\citep{chen2024spin} improves models through self-play and iterative fine-tuning; RISE~\citep{li2026rise} shows that task-induced policy changes can become self-teacher supervision; and Recursive OPD~\citep{yin2026recursiveopd} studies recursive training with dynamically co-evolving teachers and students. These works establish that accumulated experience can serve not only immediate parameter updates but also later supervision. Video reasoning, however, additionally requires joint reasoning over entities, actions, and spatiotemporal evidence, raising the question of how recursive learning should accommodate video-specific supervision. Existing video-oriented RSI studies mainly evolve systems around the model: VideoHarness-RSI~\citep{xu2026videoharness} recursively improves video context construction; Video-RSI~\citep{luo2026videorsi} uses active video investigation to improve evidence acquisition and execution; and VideoHarness-SRSI~\citep{lin2026videoharnesssrsi} emphasizes reliable evidence use and result verification during recursive search. Their optimization targets are primarily external harnesses rather than VideoLLM parameters. How to formulate video RSI as an iterative parameter-training process therefore remains an important direction. In contrast to these harness-based approaches, RECAST explicitly casts RSI as a training-based loop that updates VideoLLM parameters through stagewise reward learning and self-teacher renewal. Its \emph{Outcome-Grounded Decision Refinement} additionally evaluates choices at key teacher--student disagreements using student continuation rewards, linking evolving supervision with local decision optimization.


\section{Dataset: RECAST-100k}
\label{sec:dataset}

RECAST learns from task rewards and from a self-teacher that uses answer and spatiotemporal evidence. We construct \emph{RECAST-100k} to provide these forms of supervision in a shared training format. Starting from the STGR-RL data of Open-o3-Video~\citep{meng2026openo3}, we retain the original records and add general video questions, questions with visual evidence, and temporal localization examples.

\subsection{Task Coverage and Supervision}
\label{sec:dataset-base}

The original STGR-RL snapshot contains 37,231 training records covering image QA, general video QA, temporal localization, and questions with spatial evidence. We extend it by 62,769 records to form a collection of 100,000 examples. Of the added records, 11,416 provide general video QA, 30,233 connect visual answers to a selected frame and visible objects, and 21,120 provide temporal localization. The final collection contains 95,000 video-question records and 5,000 image-question records. Multiple questions can refer to different events in the same training video.

These tasks provide different evidence for the same training process. General questions supervise the final answer. Temporal examples identify the interval in which an event occurs. Grounded examples link an answer to a key frame and object boxes. In Sec.~\ref{sec:method}, the answers and annotations define the reward targets $a_i$ and the available evidence views $E_i^m$. The student answers from the visual input and question, while the training annotations guide reward evaluation and self-teacher scoring.

The task and source composition is shown in Fig.~\ref{fig:video-rsi-overview}; detailed counts are given in the supplementary material.


\subsection{Data Construction and Verification}
\label{sec:dataset-construction}

We first expand general video QA using LLaVA-Video-178K~\citep{zhang2024videoinstruction} and the HumanQA and LongCapQA sources of Molmo2~\citep{clark2026molmo2}. A multimodal teacher reads the video and generates multiple-choice or short-answer questions from visible content. The same teacher then answers each question without seeing its generated answer, and we retain questions that pass the answer-consistency and visual-support checks. After filtering identifiable overlap and separating held-out videos, this stage adds 11,416 training examples.

We next build questions with spatial evidence from the event queries and time intervals of TimeLens-100K~\citep{zhang2026timelens}. The event interval guides frame selection, but the teacher must generate a question supported by a visible frame. It selects one frame and identifies relevant objects, which are localized with bounding boxes. A separate verification pass checks the question, answer, chosen frame, and boxes together. The resulting 30,233 training examples use the same single-key-frame evidence format as the original grounded data.

Finally, we add temporal localization examples from the remaining TimeLens training events. Rather than generate new boundaries, we retain the existing event queries and time intervals after checking the media and annotations. We select 21,120 records across source datasets and video groups. The combined collection reaches 100,000 training records.

The three stages produce answer labels, temporal intervals, and spatial evidence in the original STGR-RL task schema. We keep their source identities and train or held-out assignments throughout construction, and exclude identified overlap with protected evaluation videos. Construction details and prompts appear in Appendices~\ref{app:dataset} and~\ref{app:data-construction-prompts}.
\FloatBarrier
\section{RECAST}
\label{sec:method}

RECAST trains one VideoLLM, called the \emph{student}. Given a video question $x_i$, the policy $\pi_\theta$ generates $y_i$; training annotations $a_i$ determine the task reward $R(x_i,y_i,a_i)$ for answer, temporal, spatial, and formatting criteria, but are not available during student generation. The annotations come from the task fields described in Sec.~\ref{sec:dataset}. When present, event intervals, selected key frames, and object boxes also form the temporal and spatial evidence views used below.

Each round alternates two stages. \emph{Reward learning} moves the student from $\theta_r^-$ to $\theta_r^+$; their frozen snapshots and video evidence define a renewed \emph{self-teacher}. During \emph{teacher-guided learning}, \emph{Future-Aware Semantic Credit} uses current and later semantic chunks to reweight advantages on sampled tokens, while \emph{Outcome-Grounded Decision Refinement} evaluates alternative tokens at one disagreement-selected position through reward-scored continuations. Both signals enter a single loss that updates the same student. The resulting $\theta_{r+1}^-$ starts the next round. The full cycle is summarized in Fig.~\ref{fig:video-rsi-overview}.

\subsection{Policy-Extrapolated Self-Teacher}
\label{sec:stage}

Reward learning changes the student's token preferences. Instead of copying only its final checkpoint, we also extrapolate the \emph{direction} of this change using two frozen snapshots. This captures recent reward-driven progress without independently training a teacher network.

Let $\theta_r^-$ and $\theta_r^+$ be the pre- and post-reward student parameters. At position $t$ of response $i$, $h_{i,t}=(x_i,y_{i,<t})$ is the video question and generated prefix. To isolate the stage-wise policy change, both snapshots predict candidate token $v$ under the same prefix and evidence view $E_i^m$, where $m\in\mathcal M_i$ denotes answer, answer--temporal, or answer--spatial evidence. Write $p_{r,i,t}^{m,\pm}(v)=\pi_{\theta_r^\pm}(v\mid h_{i,t},E_i^m)$. We extrapolate their log-probability change:
\begin{align}
\Delta_{r,i,t}^m(v)
 &=\operatorname{clip}\!\left(\alpha\log\frac{p_{r,i,t}^{m,+}(v)}{p_{r,i,t}^{m,-}(v)},-c_\Delta,c_\Delta\right),\nonumber\\
q_{r,i,t}^m(v)
 &=\operatorname{softmax}_{v}\!\left(\log p_{r,i,t}^{m,+}(v)+\Delta_{r,i,t}^m(v)\right).
\label{eq:teacher}
\end{align}
Here $\alpha$ scales the policy change and $c_\Delta$ bounds it; $q_{r,i,t}^m(v)$ is a next-token probability under view $m$. Averaging the separately normalized view distributions gives
\begin{equation}
q_{r,i,t}^{\mathrm{stage}}(v)=\frac{1}{|\mathcal M_i|}
\sum_{m\in\mathcal M_i}q_{r,i,t}^m(v).
\label{eq:teacher-fusion}
\end{equation}
This \emph{self-teacher distribution} is recomputed on each new student prefix using frozen snapshots; it is renewed from new snapshots only after the next reward-learning stage.

\subsection{Structured Teacher-Guided Optimization}
\label{sec:branch}

The renewed teacher provides two signals: one reweights \emph{sampled tokens}, while the other evaluates \emph{unselected alternatives}. Neither treats teacher probabilities as ground truth.
 Figure~\ref{fig:video-rsi-guided} illustrates the two paths; their outputs contribute to one Student update, not two sequential updates.

\begin{figure}[t]
    \centering
    \includegraphics[width=\textwidth]{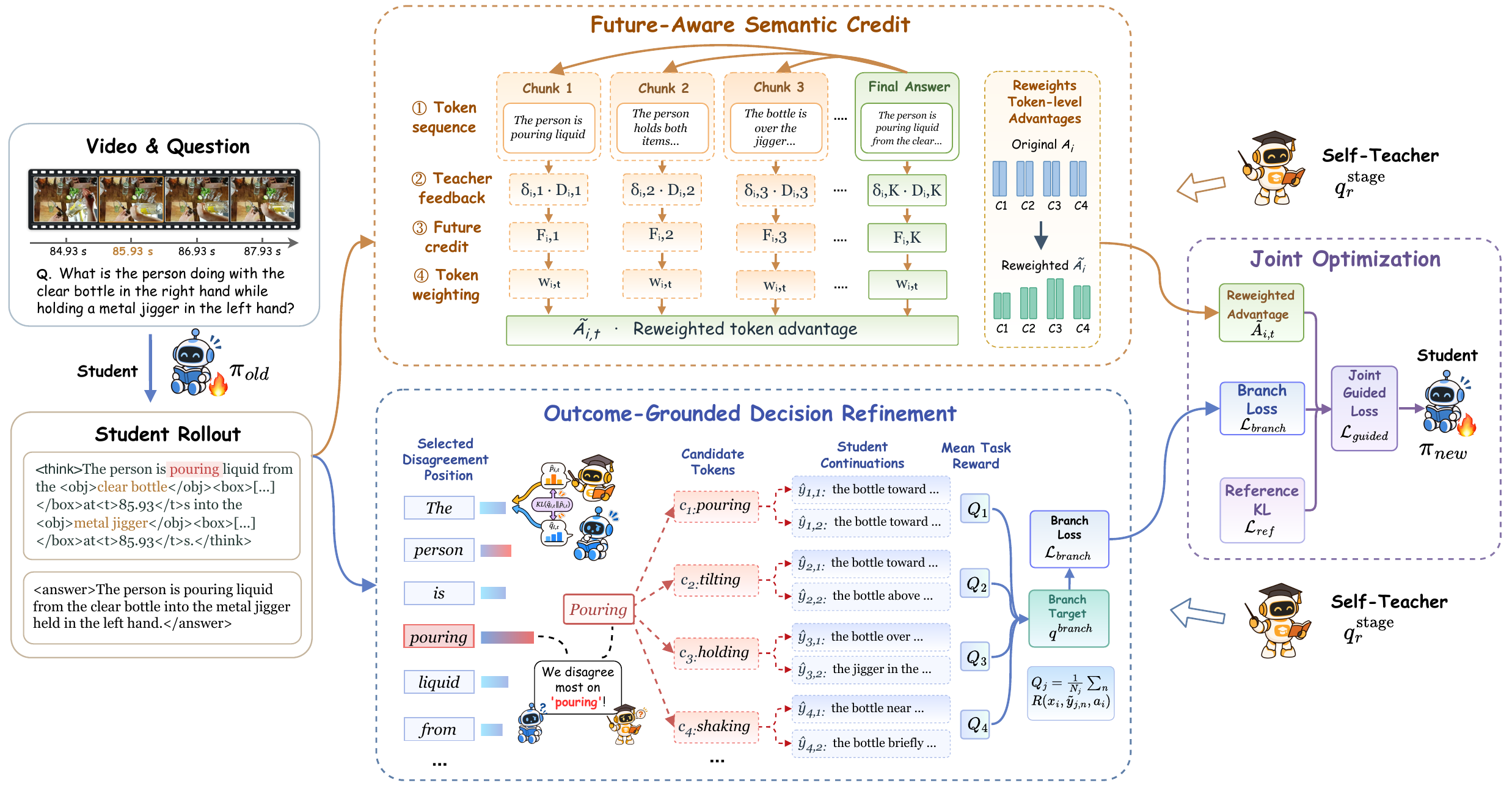}
    \caption{\textbf{Structured teacher-guided optimization.} Future-Aware Semantic Credit reweights token advantages, while Outcome-Grounded Decision Refinement uses continuation rewards to form local targets. Both guide one update.}
    \label{fig:video-rsi-guided}
\end{figure}

\noindent\textbf{Future-Aware Semantic Credit.}
This mechanism redistributes the response-level \emph{advantage} across sampled tokens without changing the reward or its sign. For example, ``picks up the red cup on the right at 00:04'' binds an entity, action, time, and location. These related fields are kept together where possible in consecutive, nonoverlapping response chunks $C_{i,1},\ldots,C_{i,K_i}$. Other text follows sentence or clause boundaries, and the final answer forms a separate chunk. Chunks segment the \emph{response}, not video frames.

We quantify the teacher's relative support for each sampled token against the pre-update student $\pi_{\mathrm{old}}$ and average within chunks ($\operatorname{sg}$ stops gradients):
\begin{align}
\delta_{i,t}&=\operatorname{sg}\!\left[\log
\frac{q_{r,i,t}^{\mathrm{stage}}(y_{i,t})}
{\pi_{\mathrm{old}}(y_{i,t}\mid h_{i,t})}\right],\nonumber\\
D_{i,k}&=\frac{1}{|C_{i,k}|}\sum_{t\in C_{i,k}}\delta_{i,t}.
\label{eq:chunk-feedback}
\end{align}
This score expresses how strongly the teacher supports the selected tokens relative to the student; it is not a new task reward.

Because early observations may matter to later reasoning, we aggregate feedback from up to $H$ later chunks \emph{in the same completed response}, discounted by $\gamma$:
\begin{equation}
F_{i,k}=\frac{\sum_{j=k}^{\min(K_i,k+H)}\gamma^{j-k}D_{i,j}}{\sum_{j=k}^{\min(K_i,k+H)}\gamma^{j-k}}.
\label{eq:future}
\end{equation}
Here $F_{i,k}$ uses the student's already completed response, so it reflects later reasoning without generating any additional continuation.

We use $F_{i,k}$ to reweight the reward-derived group advantage $A_i$. Let $k(t)$ be token $t$'s chunk and $\overline F_i$ the token-count-weighted mean credit:
\begin{align}
w_{i,t}&=\operatorname{clip}\!\left(\exp\!\left[\operatorname{sign}(A_i)\big(F_{i,k(t)}-\overline F_i\big)\right],1-c_w,1+c_w\right),\nonumber\\
\widetilde A_{i,t}&=\operatorname{sg}(A_iw_{i,t}).
\label{eq:credit}
\end{align}
Since $w_{i,t}>0$, the reward-derived advantage sign is preserved: high-credit chunks are reinforced more when $A_i>0$, and low-credit chunks are penalized more when $A_i<0$. Thus, task reward still determines the sign of the policy update and teacher feedback only adjusts its strength. Invalid feedback retains $A_i$.

\noindent\textbf{Outcome-Grounded Decision Refinement.}
Semantic credit reweights \emph{chosen} tokens but cannot assess \emph{unchosen} alternatives. We therefore test replacements at a teacher--student disagreement position and use completed outcomes, rather than teacher preference, to judge them. A large disagreement marks an informative location, not necessarily a better teacher-selected token.

Let $\mathcal C_{i,t}$ be the student's top-$k$ candidates, and $\mathcal B_{\mathcal C}$ map a distribution into those candidate probabilities plus one residual-vocabulary bucket. For $\widehat q_{i,t}=\mathcal B_{\mathcal C_{i,t}}(q_{r,i,t}^{\mathrm{stage}})$ and $\widehat p_{i,t}=\mathcal B_{\mathcal C_{i,t}}(\pi_{\mathrm{old}}(\cdot\mid h_{i,t}))$, select
\begin{align}
u_{i,t}&=\operatorname{KL}(\widehat q_{i,t}\Vert\widehat p_{i,t}),\nonumber\\
t_i^*&=\underset{t\in\mathcal E_i}{\operatorname{argmax}}\ u_{i,t}.
\label{eq:disagreement}
\end{align}
The eligible set $\mathcal E_i$ requires valid teacher feedback, sufficient candidate coverage, minimum disagreement, and continuation budget. Thus \emph{one} position is selected per response, not per chunk; an empty eligible set skips this branch.

Holding the video and prefix $y_{i,<t_i^*}$ fixed, we force each evaluated top-$k$ candidate $c_j$ and sample fresh suffixes $z_{j,n}$ from the old student, forming $\widetilde y_{j,n}=y_{i,<t_i^*}\oplus c_j\oplus z_{j,n}$. The \emph{original task reward} estimates each candidate's value:
\begin{equation}
Q_j=\frac{1}{N_j}\sum_{n=1}^{N_j} R(x_i,\widetilde y_{j,n},a_i).
\label{eq:values}
\end{equation}
Only valid new continuations contribute; terminal candidates are scored directly. Thus $Q_j$ measures student continuation outcomes, not teacher preference, and the original response reward is not mixed into the candidate average.

We adjust the old distribution $p(v)=\pi_{\mathrm{old}}(v\mid h_{i,t_i^*})$ rather than selecting a single winning token. For evaluated candidates $\mathcal J$, define coverage $\omega=\sum_{j\in\mathcal J}p(c_j)$ and reward baseline $Q_0=\sum_{j\in\mathcal J}p(c_j)Q_j/\omega$. Use the bounded bias $b(c_j)=\kappa\omega\operatorname{clip}(Q_j-Q_0,-c_Q,c_Q)$ for evaluated candidates and $b(v)=0$ otherwise:
\begin{equation}
q_i^{\mathrm{branch}}(v)=\frac{p(v)e^{b(v)}}{\sum_{u\in\mathcal V}p(u)e^{b(u)}}.
\label{eq:target}
\end{equation}
The target increases relative support for higher-reward candidates while remaining anchored to $p$, instead of forcing the student to match an unverified teacher distribution. With at least two valid candidates, forward KL to the current student on the same candidate-plus-residual buckets gives $\mathcal L_{\mathrm{branch}}$; otherwise it is skipped. \emph{The teacher selects where to test; rewards determine candidate preferences}.

\noindent\textbf{Joint Optimization.}
The two signals update one student without changing the original reward: credit reweights sampled-token advantages, and decision refinement supplies a local target. For reward group $g(i)$, let $A_i=\operatorname{sg}[(R_i-\overline R_{g(i)})/(s_{g(i)}+\varepsilon_A)]$, with group mean $\overline R_{g(i)}$ and sample standard deviation $s_{g(i)}$. Let $\mathcal L_{\mathrm{pg}}(B)$ be the clipped group-relative policy loss~\citep{shao2024deepseekmath} and $\mathcal L_{\mathrm{ref}}$ the reference-policy KL. The stages optimize:
\begin{align}
\mathcal L_{\mathrm{task}}
 &=\mathcal L_{\mathrm{pg}}(A)+\lambda_{\mathrm{ref}}\mathcal L_{\mathrm{ref}},\nonumber\\
\mathcal L_{\mathrm{guided}}
 &=\mathcal L_{\mathrm{pg}}(\widetilde A)+\lambda_{\mathrm{ref}}\mathcal L_{\mathrm{ref}}
 +\lambda_B\mathcal L_{\mathrm{branch}}.
\label{eq:objectives}
\end{align}
Snapshots, the old sampling policy, rewards, credit weights, and branch targets are detached. Only the current student receives gradients; new responses refresh step-level feedback, while the next reward stage refreshes the teacher. Objectives and mathematical properties appear in Appendix~\ref{app:training-details}.

\begin{table}[t]
\setlength{\belowcaptionskip}{5pt}
\centering
\caption{\textbf{Video reasoning and grounding results (\%).} Avg. is the unweighted mean of WorldSense Overall, VideoMMMU Overall, LRR accuracy, and TVGBench mIoU. Bold and underlined values mark the best and second-best open-source results; GPT-4o is included for reference. $\Delta$ denotes gains over Qwen2.5-VL-7B in percentage points; ``--'' indicates unavailable results.}
\label{tab:general}

\small
\renewcommand{\arraystretch}{1.05}
\setlength{\tabcolsep}{5pt}

\begin{adjustbox}{max width=\linewidth}
\begin{NiceTabular}{@{}lccccccc@{}}
\CodeBefore
\rowcolor{resultAzure}{10}
\rowcolor{deltaApricot}{11}
\Body
\toprule
\multirow{2}{*}[-2.7pt]{\textbf{Model}}
& \multicolumn{2}{c}{\textbf{WorldSense}}
& \multicolumn{2}{c}{\textbf{VideoMMMU}}
& \textbf{LRR}
& \textbf{TVGBench}
& \multirow{2}{*}[-2.7pt]{\textbf{Avg}} \\
\cmidrule(lr){2-3}
\cmidrule(lr){4-5}
\cmidrule(lr){6-6}
\cmidrule(lr){7-7}
& \textbf{Overall}
& \textbf{Understanding}
& \textbf{Overall}
& \textbf{Perception}
& \textbf{Acc.}
& \textbf{mIoU}
& \\
\midrule

GPT-4o
& 42.6 & --
& 61.2 & 66.0
& -- & -- & -- \\
\midrule

Qwen2.5-VL-7B~\citep{bai2025qwen}
& 35.5 & 38.4
& 51.2 & 64.7
& 59.3 & 16.3 & 40.6 \\

Video-R1-7B~\citep{feng2025videor1}
& 36.9 & 37.7
& 52.4 & 65.3
& 68.9 & 9.6 & 42.0 \\

VideoRFT-7B~\citep{wang2025videorft}
& 38.4 & 40.0
& 51.1 & 66.0
& 69.4 & 14.3 & 43.3 \\

Open-o3-Video-7B~\citep{meng2026openo3}
& 38.9 & 39.7
& 52.6 & 67.4
& 69.1 & 19.2 & 45.0 \\

VisionCoach-7B~\citep{lee2026visioncoach}
& 39.2 & 40.2
& \underline{53.8} & \underline{70.3}
& 70.7 & 21.0 & 46.2 \\

VISD-7B~\citep{lin2026visd}
& \underline{39.9} & \underline{40.4}
& \underline{53.8} & 69.3
& \underline{73.5} & \underline{24.8}
& \underline{48.0} \\
\midrule

\textbf{RECAST-7B (Ours)}
& \textbf{41.7} & \textbf{42.6}
& \textbf{57.1} & \textbf{73.7}
& \textbf{75.0} & \textbf{29.4}
& \textbf{50.8} \\

$\Delta$ over Qwen2.5-VL-7B
& +6.2 & +4.2
& +5.9 & +9.0
& +15.7 & +13.1 & +10.2 \\

\bottomrule
\end{NiceTabular}
\end{adjustbox}
\end{table}

\section{Experiments}
\label{sec:experiments}

\subsection{Experimental Settings}
\label{sec:settings}

\noindent\textbf{Implementation Details.}
We initialize RECAST-7B from Open-o3-Video-SFT-7B and train on RECAST-100k using 32 A800 GPUs, with vLLM for rollout generation. Each rollout batch contains 64 prompts with four responses each; optimizer updates use 16 prompt groups. We alternate 12 task-learning and 12 teacher-guided rollout steps per round for one epoch, using AdamW with learning rate $10^{-6}$ and cosine decay. The response budget is 1,024 tokens. We use $\alpha=0.5$, $H=2$, $\gamma=0.8$, and reference-KL coefficient $0.04$; local refinement uses four candidates and two continuations per evaluated nonterminal candidate. Further details are provided in the supplementary material. Training prompts and hyperparameters are provided in Appendices~\ref{app:training-prompts} and~\ref{app:hyperparameters}.

\noindent\textbf{Benchmarks and Metrics.}
We evaluate video reasoning on WorldSense~\citep{hong2026worldsense}, VideoMMMU~\citep{hu2026videommmu}, and LongVideo-Reason-eval (LRR)~\citep{chen2025longvila-r1} using accuracy, and temporal grounding on TVGBench~\citep{wang2025timer1} and Charades-STA~\citep{gao2017tall} using mIoU. Charades-STA additionally reports R@0.3, R@0.5, and R@0.7. General video understanding is measured by accuracy on MVBench~\citep{li2024mvbench}, VideoMME~\citep{fu2025videomme}, TempCompass~\citep{liu2024tempcompass}, and SEED-Bench~\citep{li2023seed}. Average-score definitions are given in the table captions. Protocols and baseline details appear in Appendix~\ref{app:evaluation-details}.


\begin{table}[t]
\centering
\caption{\textbf{Charades-STA temporal grounding results (\%).}}
\label{tab:charades_sta}

\small
\setlength{\tabcolsep}{4pt}
\renewcommand{\arraystretch}{1.05}

\begin{adjustbox}{max width=\linewidth}
\begin{tabular}{lcccc}
\toprule
\textbf{Model}
& \textbf{R@0.3}
& \textbf{R@0.5}
& \textbf{R@0.7}
& \textbf{mIoU} \\

\specialrule{\lightrulewidth}{\aboverulesep}{0pt}
\rowcolor{sectionMist}
\multicolumn{5}{l}{\textit{General Video LLMs}} \\
VideoChat~\citep{li2023videochat}
& 9.0 & 3.3 & 1.3 & 6.5 \\
VideoLLaMA~\citep{zhang2023videollama}
& 10.4 & 3.8 & 0.9 & 7.1 \\
Video-ChatGPT~\citep{maaz2024videochatgpt}
& 20.0 & 7.7 & 1.7 & 13.7 \\
Valley~\citep{luo2023valley}
& 28.4 & 1.8 & 0.3 & 21.4 \\

\rowcolor{sectionMist}
\multicolumn{5}{l}{\textit{Temporal Grounding Video LLMs}} \\
VTimeLLM~\citep{huang2024vtimellm}
& 51.0 & 27.5 & 11.4 & 31.2 \\
TimeChat~\citep{ren2024timechat}
& -- & 32.2 & 13.4 & -- \\
HawkEye~\citep{wang2024hawkeye}
& 50.6 & 31.4 & 14.5 & 33.7 \\
VTG-LLM~\citep{guo2025vtgllm}
& 51.2 & 33.8 & 15.7 & 34.4 \\

\rowcolor{sectionMist}
\multicolumn{5}{l}{\textit{Grounded Reasoning Models}} \\
Open-o3-video~\citep{meng2026openo3}
& 62.6 & 45.6 & \underline{24.5} & 42.5 \\
VISD~\citep{lin2026visd}
& \underline{67.3} & \underline{46.1}
& 21.6 & \underline{44.6} \\

\rowcolor{resultAzure}
\textbf{RECAST-7B (Ours)}
& \textbf{74.7} & \textbf{55.7}
& \textbf{31.0} & \textbf{50.6} \\

\specialrule{\heavyrulewidth}{0pt}{0pt}
\end{tabular}
\end{adjustbox}
\end{table}

\subsection{Main Results}
\noindent\textbf{Video Reasoning and Grounding.}
RECAST leads all four benchmarks among the compared open-source methods (Table~\ref{tab:general}), achieving a 50.8\% average score, 2.8 percentage points above VISD. It exceeds VISD by 3.3 percentage points in VideoMMMU accuracy and 4.6 in TVGBench mIoU. These gains show that RECAST improves both video reasoning and temporal evidence localization.

\noindent\textbf{Temporal Grounding on Charades-STA.}
RECAST leads all four metrics (Table~\ref{tab:charades_sta}), achieving 55.7\% R@0.5 and 50.6\% mIoU. These exceed VISD by 9.6 and 6.0 percentage points, respectively. The joint gains in recall and temporal overlap indicate more accurate event localization.

\noindent\textbf{General Video Understanding.}
RECAST leads all four benchmarks (Table~\ref{tab:general_video_understanding}), with 65.0\% average accuracy, 2.1 percentage points above VideoRFT. On SEED-Bench, it surpasses TW-GRPO by 3.1 percentage points. These gains support its effectiveness in general video understanding beyond specialized grounding tasks. Qualitative examples and failure cases appear in Appendix~\ref{app:cases}.



\begin{table}[t]
\centering
\caption{\textbf{General video understanding results (\%).} Avg.: four-benchmark mean; $\Delta$: gains in percentage points.}
\label{tab:general_video_understanding}

\footnotesize
\setlength{\tabcolsep}{2.5pt}
\renewcommand{\arraystretch}{1.10}

\begin{adjustbox}{max width=\columnwidth}
\begin{NiceTabular}{@{}lccccc@{}}
\CodeBefore
\rowcolor{resultAzure}{9}
\rowcolor{deltaApricot}{10}
\Body
\toprule
\textbf{Model}
& \textbf{MVBench}
& \textbf{VideoMME}
& \textbf{TempCompass}
& \textbf{SEED-Bench} & \textbf{Avg.} \\
\midrule
LLAVA-OV-7B~\citep{li2024llavaonevision}
& 56.7 & 58.2 & 67.2 & -- & --\\
Qwen2.5-VL-7B~\citep{bai2025qwen}
& 60.8 & 56.5 & 68.6 & 57.2 & 60.8\\
Video-R1~\citep{feng2025videor1}
& 65.5 & 57.2 & 68.5 & 57.5 & 62.2 \\
VideoChat-R1~\citep{li2025videochatr1}
& \underline{66.0} & 54.7 & \underline{69.8} & 57.5 & 62.0\\
VideoRFT~\citep{wang2025videorft}
& 65.3 & \underline{59.2} & 69.4 & 57.8 & \underline{62.9}\\
TW-GRPO~\citep{dang2025twgrpo}
& 65.6 & 56.6 & 68.9 & \underline{57.9} & 62.3 \\
Open-o3-video~\citep{meng2026openo3}
& 58.2 & 51.2 & 64.4 & 55.9 & 57.4\\

\midrule
\textbf{RECAST-7B (Ours)}
& \textbf{66.9} & \textbf{60.8}
& \textbf{71.3} & \textbf{61.0} & \textbf{65.0}\\
$\Delta$ over Qwen2.5-VL-7B & +6.1 & +4.3 & +2.7 & +3.8 & +4.2 \\
\bottomrule
\end{NiceTabular}
\end{adjustbox}
\end{table}


\subsection{Ablation Studies}

We ablate Future-Aware Semantic Credit (FASC), Outcome-Grounded Decision Refinement (ODR), and key design choices (Tables~\ref{tab:ablation_fasc_odr} and~\ref{tab:component_ablation}). We further ablate training and feedback settings and analyze model scaling and recursive improvement (Figures~\ref{fig:hyperparameter_sensitivity}--\ref{fig:model_scaling}, Figure~\ref{fig:video-rsi-motivation}). Ablation settings appear in Appendix~\ref{app:additional-experiments}.

\begin{figure}[htbp]
    \centering
    \includegraphics[width=\textwidth]{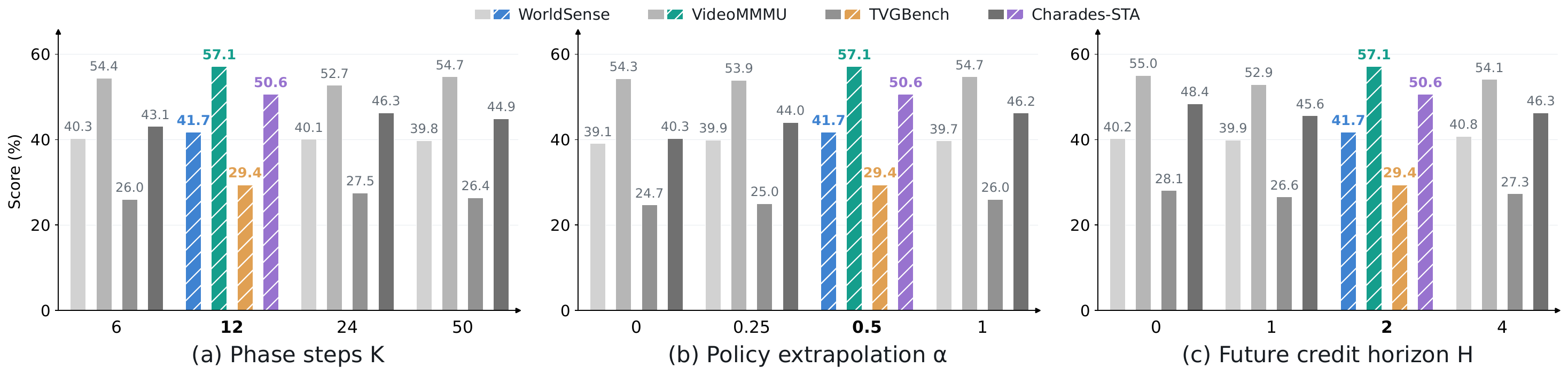}
\caption{\textbf{Training and feedback ablations.} Effects of (a) phase length $K$, (b) extrapolation coefficient $\alpha$, and (c) future credit horizon $H$. Hatched bars mark the defaults ($K=12$, $\alpha=0.5$, $H=2$), which perform best across all four benchmarks among the tested settings.}
    \label{fig:hyperparameter_sensitivity}
\end{figure}

\begin{figure}[htbp]
    \centering
    \includegraphics[width=\textwidth]{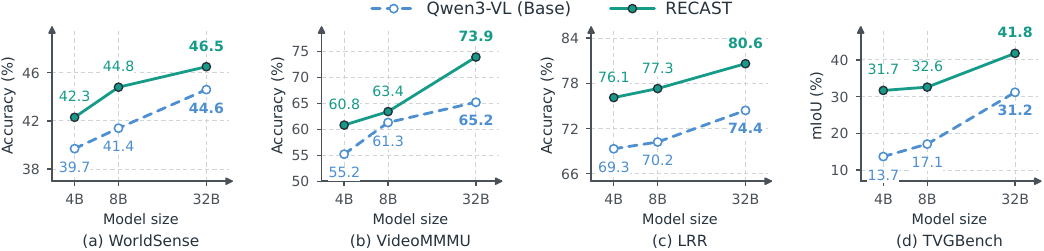}
    \caption{\textbf{Performance across model scales.} RECAST improves over Qwen3-VL baselines at 4B, 8B, and 32B on WorldSense, VideoMMMU, and LRR (accuracy), and TVGBench (mIoU).}
    \label{fig:model_scaling}
\end{figure}

\begin{table}[t]
    \centering
\caption{\textbf{FASC and ODR ablations (\%).} Metrics: overall accuracy (WorldSense, VideoMMMU), accuracy (LRR), and mIoU (TVGBench, Charades-STA). Gray superscripts show drops from the full model in percentage points.}
    \label{tab:ablation_fasc_odr}
    \setlength{\tabcolsep}{3pt}
    \begin{adjustbox}{max width=\linewidth}
        \begin{tabular}{ccccccc}
            \toprule
            \textbf{FASC} & \textbf{ODR}
            & \textbf{WorldSense}
            & \textbf{VideoMMMU}
            & \textbf{LRR}
            & \textbf{TVGBench}
            & \textbf{Charades-STA} \\
            \midrule

            \ding{55} & \ding{55}
            & \abldrop{39.3}{2.4}
            & \abldrop{53.9}{3.2}
            & \abldrop{71.3}{3.7}
            & \abldrop{19.0}{10.4}
            & \abldrop{31.2}{19.4} \\

            \ding{51} & \ding{55}
            & \abldrop{40.2}{1.5}
            & \abldrop{54.7}{2.4}
            & \abldrop{72.9}{2.1}
            & \abldrop{23.1}{6.3}
            & \abldrop{39.5}{11.1} \\

            \ding{55} & \ding{51}
            & \abldrop{39.6}{2.1}
            & \abldrop{55.1}{2.0}
            & \abldrop{72.4}{2.6}
            & \abldrop{23.1}{6.3}
            & \abldrop{38.4}{12.2} \\

            \specialrule{\lightrulewidth}{\aboverulesep}{0pt}
            \rowcolor{ourblue}
            \ding{51} & \ding{51}
            & \textbf{41.7}
            & \textbf{57.1}
            & \textbf{75.0}
            & \textbf{29.4}
            & \textbf{50.6} \\
            \specialrule{\heavyrulewidth}{0pt}{0pt}
        \end{tabular}
    \end{adjustbox}
\end{table}

\begin{table}[t]
    \centering
    \caption{\textbf{RECAST design ablations.} PE: policy extrapolation; TR: teacher renewal; FC: future credit; NSC: non-semantic chunks; TLT: teacher local targets. Notation follows Table~\ref{tab:ablation_fasc_odr}.}
    \label{tab:component_ablation}
    \setlength{\tabcolsep}{4pt}
    \begin{adjustbox}{max width=\linewidth}
        \begin{tabular}{lccccc}
            \toprule
            \textbf{Variant}
            & \textbf{WorldSense}
            & \textbf{VideoMMMU}
            & \textbf{LRR}
            & \textbf{TVGBench}
            & \textbf{Charades-STA} \\
            \midrule

            w/o PE
            & \abldrop{39.1}{2.6}
            & \abldrop{54.3}{2.8}
            & \abldrop{71.9}{3.1}
            & \abldrop{24.7}{4.7}
            & \abldrop{40.3}{10.3} \\

            w/o TR
            & \abldrop{40.7}{1.0}
            & \abldrop{56.2}{0.9}
            & \abldrop{73.4}{1.6}
            & \abldrop{26.6}{2.8}
            & \abldrop{42.1}{8.5} \\

            w/o FC
            & \abldrop{40.2}{1.5}
            & \abldrop{55.0}{2.1}
            & \abldrop{73.5}{1.5}
            & \abldrop{28.1}{1.3}
            & \abldrop{48.4}{2.2} \\

            w/ NSC
            & \abldrop{38.4}{3.3}
            & \abldrop{54.7}{2.4}
            & \abldrop{71.8}{3.2}
            & \abldrop{27.8}{1.6}
            & \abldrop{49.5}{1.1} \\

            w/ TLT
            & \abldrop{39.7}{2.0}
            & \abldrop{55.3}{1.8}
            & \abldrop{72.9}{2.1}
            & \abldrop{28.2}{1.2}
            & \abldrop{48.6}{2.0} \\

            \rowcolor{ourblue}
            \textbf{Full}
            & \textbf{41.7}
            & \textbf{57.1}
            & \textbf{75.0}
            & \textbf{29.4}
            & \textbf{50.6} \\
            \specialrule{\heavyrulewidth}{0pt}{0pt}
        \end{tabular}
    \end{adjustbox}
\end{table}

\subsubsection{Component and Design Ablations}

\noindent\textbf{Effect of FASC and ODR.}
Table~\ref{tab:ablation_fasc_odr} shows that adding FASC or ODR alone increases Charades-STA mIoU from 31.2\% to 39.5\% or 38.4\%, respectively. Combining both reaches 50.6\%, exceeding the stronger single-component variant by 11.1 percentage points. This supports their complementary roles in reweighting sampled-token advantages and evaluating alternative decisions.

\noindent\textbf{Effect of Policy Extrapolation and Teacher Renewal.}
The \textit{w/o PE} variant sets $\alpha=0$, constructing the teacher from each round's post-task model while retaining renewal. The \textit{w/o TR} variant fixes the first-round teacher snapshots while student training continues. Table~\ref{tab:component_ablation} shows corresponding Charades-STA mIoU drops of 10.3 and 8.5 percentage points relative to the full model. The comparisons support both extrapolating reward-induced policy changes and refreshing the teacher across rounds.

\noindent\textbf{Effect of Semantic Credit Assignment.}
In Table~\ref{tab:component_ablation}, the \textit{w/o FC} variant sets $H=0$ and retains current-chunk feedback, lowering Charades-STA mIoU by 2.2 percentage points. The \textit{w/ NSC} variant replaces semantic chunks with fixed-length token chunks while retaining the remaining credit rules, reducing WorldSense by 3.3 percentage points. These results support both organizing feedback around related event information and incorporating feedback from later response chunks.

\noindent\textbf{Effect of Outcome-Grounded Local Targets.}
The \textit{w/ TLT} variant retains position selection, the candidate set, and the KL objective, but replaces the target constructed from candidate continuation rewards with teacher probabilities. This replacement lowers Charades-STA and TVGBench mIoU by 2.0 and 1.2 percentage points, respectively (Table~\ref{tab:component_ablation}). With the other local refinement choices held fixed, this comparison favors constructing targets from task outcomes.

\subsubsection{Ablation of Training and Feedback Settings}

Figure~\ref{fig:hyperparameter_sensitivity} favors an intermediate alternation interval: $K=12$ outperforms both shorter and longer phases. Moderate extrapolation ($\alpha=0.5$) improves Charades-STA mIoU by 4.4 percentage points over $\alpha=1$, supporting controlled amplification of reward-induced policy changes. Likewise, $H=2$ exceeds $H=4$ by 4.3 percentage points on the same metric, favoring a short window of later-chunk feedback.

\subsubsection{Base Model Ablations}

Figure~\ref{fig:model_scaling} compares Qwen3-VL base models with their RECAST-trained counterparts at 4B, 8B, and 32B parameters. RECAST improves all four benchmarks at every evaluated model size. TVGBench mIoU increases by 18.0, 15.5, and 10.6 percentage points, respectively, while VideoMMMU improves from 65.2\% to 73.9\% at 32B. These results show that the method benefits both smaller and larger models within the evaluated family, with gains extending beyond grounding to video reasoning.


\subsubsection{Recursive Improvement}

As shown in the bottom-right corner of Figure~\ref{fig:video-rsi-motivation}, RECAST improves its mean evaluation score from 46.70\% to 50.76\%, outperforming the variant without self-teacher renewal by 2.96 percentage points. The full model resumes improvement after an intermediate plateau, while the fixed-teacher variant fluctuates around 47\%--48\%. These results support teacher renewal as a key mechanism for sustained recursive improvement, turning reward-driven policy progress into updated guidance for subsequent training rounds.
\FloatBarrier

\section{Conclusion}
\label{sec:conclusion}
RECAST enables training-based recursive self-improvement in VideoLLMs through evidence-conditioned self-teacher renewal from reward-induced policy changes; Future-Aware Semantic Credit reweights advantages across semantic chunks, while Outcome-Grounded Decision Refinement refines alternatives via reward-scored student continuations. RECAST-100k provides 100,000 spatiotemporally enriched examples; RECAST-7B leads compared open-source models across nine benchmarks, reaching 50.8 in reasoning-and-grounding, 50.6 Charades-STA mIoU, and 65.0 in general-video understanding; sustained gains across rounds and scales support a broadly applicable training paradigm.

\endgroup

\bibliographystyle{plainnat}
\bibliography{main}

\clearpage
\appendix
\startcontents[appendix]
\section*{Appendix Contents}
\printcontents[appendix]{}{1}{\setcounter{tocdepth}{2}}
\clearpage
\setcounter{figure}{0}
\setcounter{table}{0}
\setcounter{equation}{0}
\renewcommand{\thefigure}{S\arabic{figure}}
\renewcommand{\thetable}{S\arabic{table}}
\renewcommand{\theequation}{S\arabic{equation}}
\renewcommand{\theHfigure}{supp.\arabic{figure}}
\renewcommand{\theHtable}{supp.\arabic{table}}
\renewcommand{\theHequation}{supp.\arabic{equation}}
\section{Prompt Templates}
\label{app:prompts}

Figures~\ref{fig:app:training-prompts}--\ref{fig:app:data-construction-prompts} show the training, evaluation, and data-construction templates. Braces mark runtime fields.

\subsection{Training Prompts}
\label{app:training-prompts}

The student receives task-specific instructions for citing evidence and formatting its answer. The self-teacher also receives available hindsight evidence in the user message.

\noindent\begin{minipage}{\linewidth}
  \centering
  \includegraphics[page=1,trim=31bp 51bp 30bp 30bp,clip,width=\linewidth,height=.70\textheight,keepaspectratio]{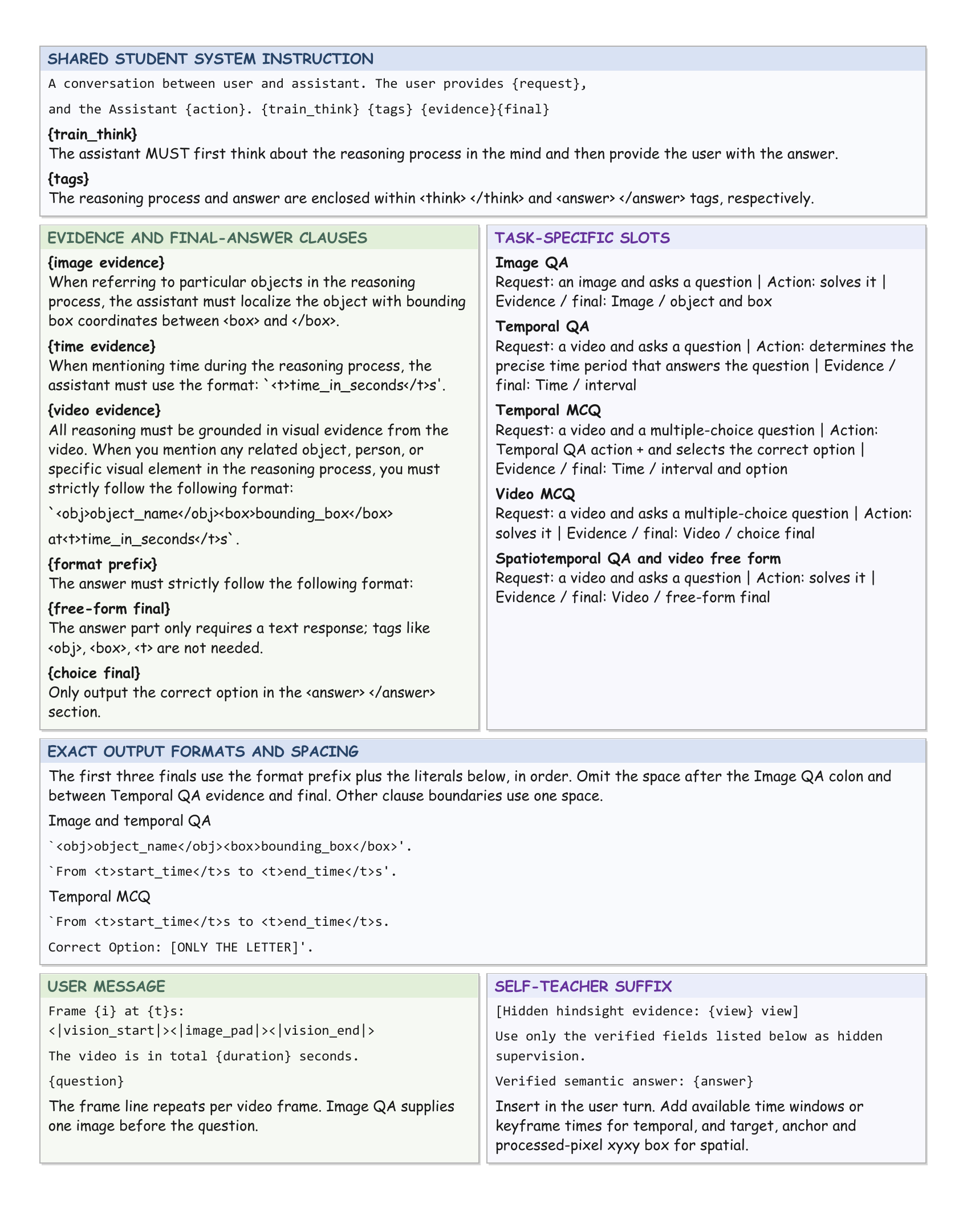}
  \captionsetup{hypcap=false}
  \captionof{figure}{\textbf{Training prompt templates.} Shared student instructions, task-specific clauses, and hindsight evidence for the self-teacher.}
  \label{fig:app:training-prompts}
\end{minipage}\par

\clearpage
\subsection{Evaluation Prompts}
\label{app:evaluation-prompts}

\begin{figure}[!htbp]
  \centering
  \includegraphics[trim=29bp 80bp 36bp 30bp,clip,width=\linewidth,height=.83\textheight,keepaspectratio]{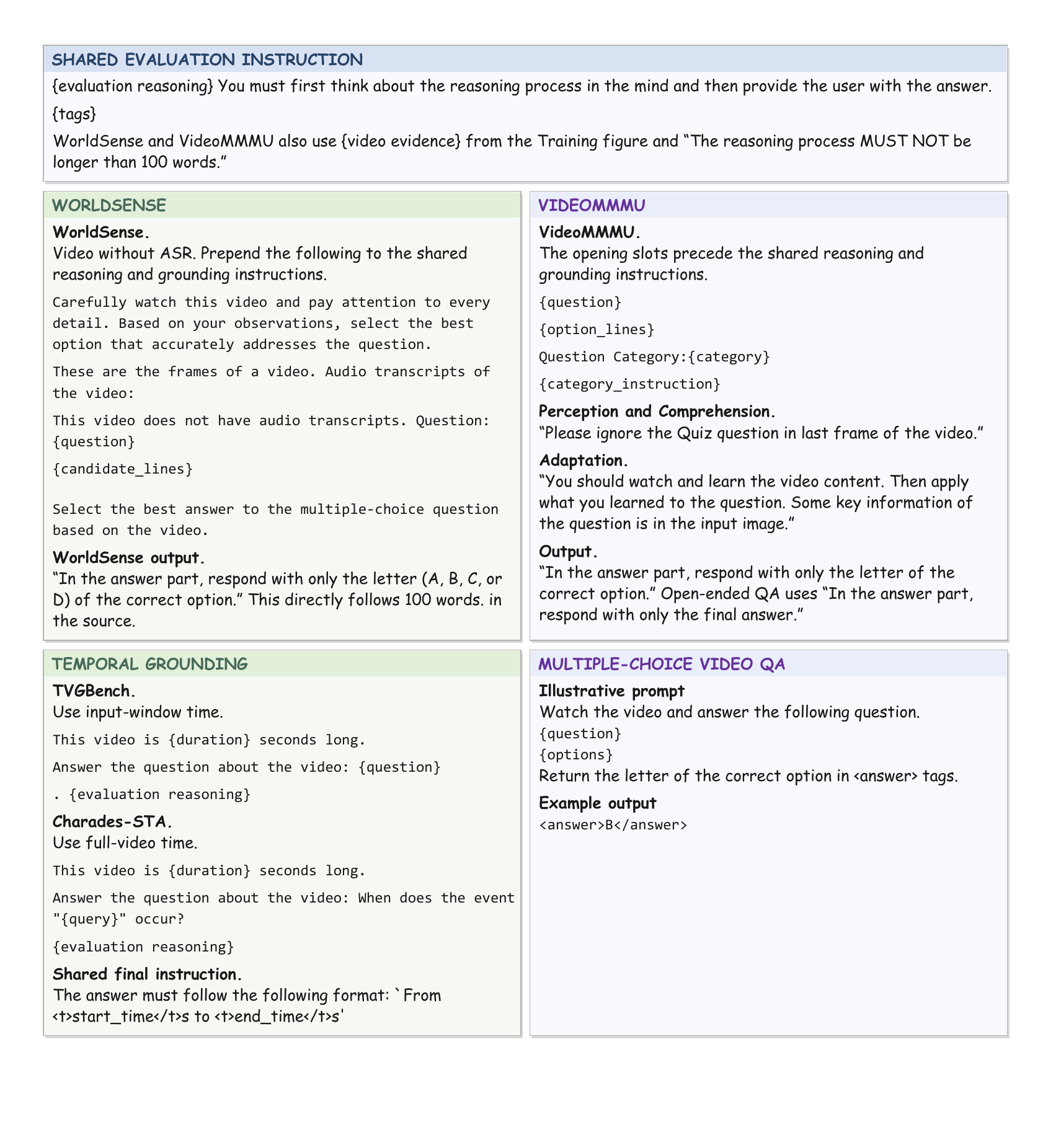}
  \caption{\textbf{Evaluation prompt templates.} The panels show the shared reasoning instruction, benchmark-specific inputs, and final-answer formats for video QA and temporal grounding.}
  \label{fig:app:evaluation-prompts}
\end{figure}

The final-answer format follows the benchmark. Video QA returns an option letter or an open-ended answer. Temporal grounding returns an interval on the input-window timeline for TVGBench and on the full-video timeline for Charades-STA. Figure~\ref{fig:app:evaluation-prompts} also illustrates the input and output format for multiple-choice video QA.

\clearpage
\subsection{Data-Construction Prompts}
\label{app:data-construction-prompts}

General video QA candidates undergo two blind verification calls, with an additional consistency review for short answers. Grounded QA selects one decisive frame, localizes objects in that frame, and reviews the answer and its evidence. The native temporal branch retains the source query and interval after rule-based screening.

\begin{figure}[H]
  \centering
  \includegraphics[page=3,trim=31bp 37bp 31bp 18bp,clip,width=\linewidth,height=.73\textheight,keepaspectratio]{figures/recast_prompts.pdf}
  \par
  \caption{\textbf{RECAST-100k data-construction prompt templates.} The panels show generation, verification, and localization instructions and their output fields.}
  \label{fig:app:data-construction-prompts}
\end{figure}

\section{Extended Related Work}

\subsection{Post-Training and Self-Distillation for VideoLLMs}

Post-training for VideoLLMs has gradually evolved from outcome-based policy optimization toward finer-grained process supervision. Video-R1~\citep{feng2025videor1} and VideoRFT~\citep{wang2025videorft} employ reinforcement learning to improve video reasoning, while Open-o3-Video~\citep{meng2026openo3} further incorporates answer correctness and spatiotemporal evidence into verifiable rewards. These approaches primarily rely on response-level or trajectory-level rewards, making it difficult to distinguish the contributions of individual intermediate decisions. To provide denser supervision, Video-OPD~\citep{li2026videoopd} applies teacher-guided token-level distillation on student-generated trajectories, while Video-OPSD~\citep{wang2026videoopsd} constructs a self-teacher using task-relevant privileged visual evidence.

VISD~\citep{lin2026visd} uses structured diagnostic feedback to guide video reasoning. MeSD~\citep{zhu2026mesd} combines teacher preferences conditioned on answer, temporal, and spatial evidence to refine token updates and correct verified failures. RECAST studies how a renewed self-teacher can guide individual decisions within a video reasoning response. Future-Aware Semantic Credit uses teacher feedback from the current and subsequent response chunks to reweight the advantages of sampled tokens. Outcome-Grounded Decision Refinement uses teacher--student disagreement to select a token position, then evaluates alternative tokens by scoring student-generated continuations with the task reward.

\subsection{Recursive Self-Improvement for Video Reasoning}

Recursive self-improvement (RSI) reuses experience accumulated in previous iterations to provide supervision for subsequent training. In general-purpose language models, RISE~\citep{li2026rise} transforms reward-induced policy changes into self-teacher supervision, and Recursive OPD~\citep{yin2026recursiveopd} studies how teachers and students co-evolve during iterative training. These studies show how progress made in one iteration can provide supervision for the next. Video reasoning raises a further question of how to construct this supervision when answers depend on temporal and spatial evidence.

In the video domain, VideoHarness-RSI~\citep{xu2026videoharness} improves long-video understanding through recursive context construction, Video-RSI~\citep{luo2026videorsi} uses active video investigation to guide evidence acquisition, and VideoHarness-SRSI~\citep{lin2026videoharnesssrsi} incorporates evidence verification into recursive search. These approaches update the external harness that gathers or organizes video evidence. RECAST updates the VideoLLM over successive training rounds. For each available evidence view, two frozen student snapshots predict the next token from the same response prefix, one before and one after reward learning. RECAST extrapolates the change in their log probabilities within each view and averages the resulting distributions to form the teacher for that round. The student updated under this teacher starts the next round of reward learning.
\section{Additional Experimental Results}
\label{app:additional-experimental-results}
We assess the annotation quality of RECAST-100k and compare RECAST-7B across four training data mixtures.

\subsection{Data Quality and Validation}
\label{app:detailed-results}
We manually inspected 3,000 randomly sampled examples from RECAST-100k. We assessed temporal and spatial evidence on examples with the corresponding annotations. Question--answer validity and visual support passed in 98.1\% and 97.6\% of examples, respectively. Overall, 2,802 examples (93.4\%) were acceptable (Table~\ref{tab:sup-human-inspection-stgr}).

\noindent\begin{minipage}{\linewidth}
\centering
\captionsetup{type=table}
\caption{\textbf{Human inspection of RECAST-100k.} The audit covers 3,000 randomly sampled examples.}
\label{tab:sup-human-inspection-stgr}
\small
\setlength{\tabcolsep}{4pt}
\renewcommand{\arraystretch}{0.98}
\begin{tabularx}{\linewidth}{>{\raggedright\arraybackslash}Xrrr}
\toprule
\textbf{Criterion} & \textbf{Inspected} & \textbf{Passed} & \textbf{Rate} \\
\midrule
Question--answer validity & 3,000 & 2,943 & 98.1\% \\
Visual support & 3,000 & 2,928 & 97.6\% \\
Accurate temporal evidence & 2,040 & 1,938 & 95.0\% \\
Accurate spatial evidence & 1,260 & 1,185 & 94.0\% \\
\specialrule{\lightrulewidth}{\aboverulesep}{0pt}
\rowcolor{recasttablehighlight}
\textbf{Overall acceptable} & \textbf{3,000} & \textbf{2,802} & \textbf{93.4\%} \\
\specialrule{\heavyrulewidth}{0pt}{0pt}
\end{tabularx}
\end{minipage}\par

\subsection{Ablation Settings and Additional Experiments}
\label{app:additional-experiments} \paragraph{Component ablations.} We test all four combinations of FASC and ODR. Disabling FASC assigns the group advantage $A_i$ to every sampled token. Disabling ODR removes the branch loss. In \emph{w/o PE}, $\alpha=0$ and the teacher is still renewed. In \emph{w/o TR}, the first round's teacher snapshots stay fixed. \emph{w/o FC} sets $H=0$, retaining current-chunk credit. \emph{w/ NSC} uses fixed-length token chunks. \emph{w/ TLT} uses teacher probabilities as local targets with the same position selection, candidates, and KL loss. Figure~4 of the main paper varies phase length $K\in\{6,12,24,50\}$, extrapolation $\alpha\in\{0,0.25,0.5,1\}$, and credit horizon $H\in\{0,1,2,4\}$. The defaults are $(K,\alpha,H)=(12,0.5,2)$.
\paragraph{Training data scale and composition.}
Starting from the original STGR-RL collection, we add general video QA, followed by grounded spatiotemporal QA and temporal localization examples. Table~\ref{tab:sup-training-mixtures} lists the additions at each stage. The 100k mixture corresponds to the full RECAST-100k training set.

\noindent\begin{minipage}{\linewidth}
\centering
\captionsetup{type=table}
\caption{\textbf{Training data mixtures.} Each expanded set builds on the preceding recipe.}
\label{tab:sup-training-mixtures}
\footnotesize
\setlength{\tabcolsep}{4pt}
\begin{tabularx}{\linewidth}{lr>{\raggedright\arraybackslash}X}
\toprule
\textbf{Set} & \textbf{Records} & \textbf{Composition} \\
\midrule
37k & 37,231 & Original STGR-RL: image QA, general video QA, temporal localization, and joint spatiotemporal QA. \\
\shortstack[l]{48.6k\\(QA56)} & 48,647 & Adds 11,416 general video QA examples (7,612 multiple-choice and 3,804 open-ended). \\
69k & 68,647 & Adds 20,000 joint spatiotemporal QA examples annotated with keyframes and object boxes. \\
\specialrule{\lightrulewidth}{\aboverulesep}{0pt}
\rowcolor{recasttablehighlight}
\textbf{100k} & \textbf{100,000} & Adds 10,233 joint spatiotemporal QA and 21,120 temporal localization QA examples. \\
\specialrule{\heavyrulewidth}{0pt}{0pt}
\end{tabularx}
\end{minipage}\par\smallskip

\noindent\begin{minipage}{\linewidth}
\centering
\captionsetup{type=table}
\caption{\textbf{Performance across training data mixtures.} We evaluate each run at the indicated checkpoint. Best scores are in bold.}
\label{tab:sup-training-scale-results}
\scriptsize
\setlength{\tabcolsep}{2pt}
\renewcommand{\arraystretch}{1.10}
\begin{tabularx}{\linewidth}{>{\raggedright\arraybackslash}Xrccccc}
\toprule
\textbf{Set} & \textbf{Step} & \shortstack{\textbf{WorldSense}\\Overall} & \shortstack{\textbf{VideoMMMU}\\Overall} & \shortstack{\textbf{LRR}\\Acc.} & \shortstack{\textbf{TVGBench}\\mIoU} & \shortstack{\textbf{Charades-STA}\\mIoU} \\
\midrule
37k   & 582   & 40.6 & 54.8 & 72.9 & 26.9 & 48.4 \\
48.6k & 761   & 40.3 & 55.8 & 73.6 & 27.8 & 49.6 \\
69k   & 1,073 & 40.7 & 54.2 & 74.0 & 28.3 & 50.4 \\
\specialrule{\lightrulewidth}{\aboverulesep}{0pt}
\rowcolor{recasttablehighlight}
\textbf{100k} & 1,563 & \textbf{41.7} & \textbf{57.1} & \textbf{75.0} & \textbf{29.4} & \textbf{50.6} \\
\specialrule{\heavyrulewidth}{0pt}{0pt}
\end{tabularx}
\end{minipage}\par\smallskip

As the training mixture expands, the mean score across the five benchmarks in Table~\ref{tab:sup-training-scale-results} rises from 48.72 at 37k to 49.42 at 48.6k, 49.52 at 69k, and 50.76 at 100k. LRR accuracy and temporal grounding mIoU on both TVGBench and Charades-STA improve at every stage. Accuracy on WorldSense and VideoMMMU fluctuates across the intermediate mixtures, but all five benchmarks reach their highest scores at 100k.

\medskip
\noindent\begin{minipage}{\linewidth}
\centering
\includegraphics[width=\linewidth]{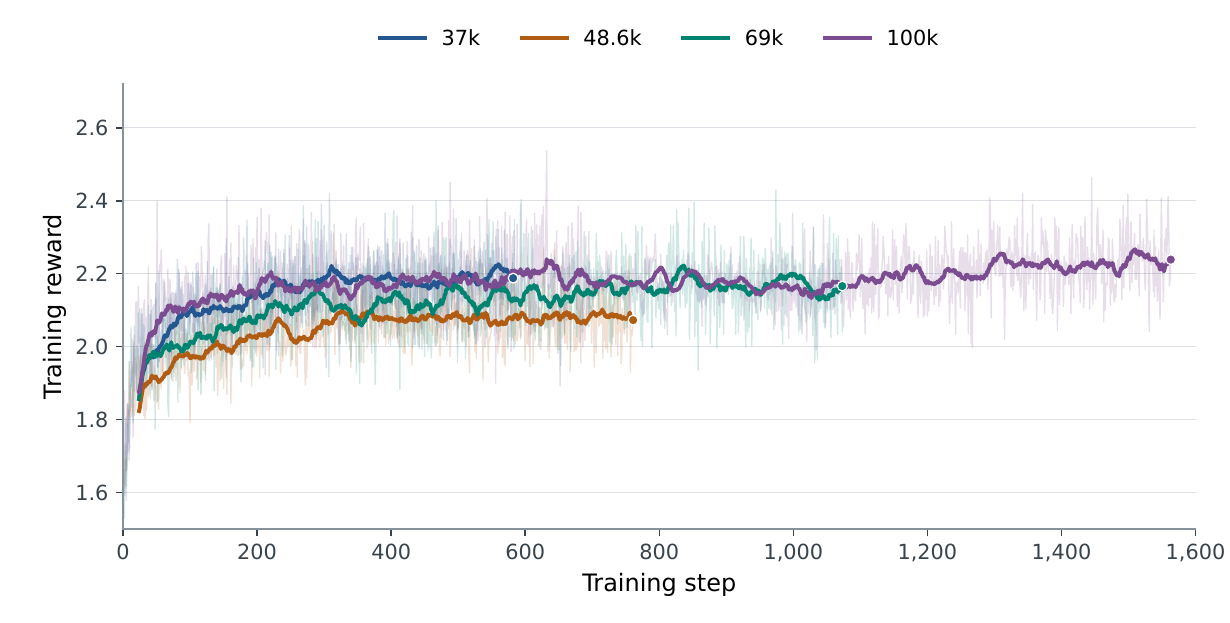}
\captionsetup{type=figure}
\caption{\textbf{Illustrative training reward curves} for the four training data mixtures.}
\label{fig:sup-training-reward-illustrative}
\end{minipage}\par\medskip

\paragraph{Training efficiency.}
Table~\ref{tab:sup-training-efficiency} reports net training time and throughput. The 100k run takes 31.92 hours, averaging 73.53 seconds per step and 277.79 tokens/s/GPU. The 69k run achieves the highest per-GPU throughput (299.22 tokens/s/GPU).

\noindent\begin{minipage}{\linewidth}
\centering
\captionsetup{type=table}
\caption{\textbf{Training time and efficiency across data mixtures.}}
\label{tab:sup-training-efficiency}
\footnotesize
\setlength{\tabcolsep}{3pt}
\renewcommand{\arraystretch}{1.05}
\begin{tabularx}{\linewidth}{>{\raggedright\arraybackslash}Xrrrr}
\toprule
\textbf{Set} & \textbf{Steps} & \shortstack{\textbf{Net time}\\(h)} & \shortstack{\textbf{Mean step time}\\(s)} & \shortstack{\textbf{Throughput}\\(tokens/s/GPU)} \\
\midrule
37k   & 582   & 10.93 & 67.61 & 280.30 \\
48.6k & 761   & 14.23 & 67.30 & 287.07 \\
69k   & 1,073 & 19.88 & 66.69 & 299.22 \\
\specialrule{\lightrulewidth}{\aboverulesep}{0pt}
\rowcolor{recasttablehighlight}
\textbf{100k} & 1,563 & 31.92 & 73.53 & 277.79 \\
\specialrule{\heavyrulewidth}{0pt}{0pt}
\end{tabularx}
\end{minipage}\par\medskip
\section{Dataset Construction and Verification}
\label{app:dataset}
RECAST-100k extends STGR-RL with general video QA, grounded spatiotemporal QA, and native temporal localization. We use Qwen3.8-27B to generate and verify the new general video QA and grounded spatiotemporal QA examples. Appendix~\ref{app:data-construction-prompts} summarizes the annotation prompts, and Appendix~\ref{app:detailed-results} reports the human audit.

\subsection{Dataset Composition}
\label{app:dataset-composition}
We retain all 37,231 records from a fixed STGR-RL training snapshot. Table~\ref{tab:app:dataset-composition} groups general video multiple-choice and free-form QA in one row. The general video QA expansion yields 48,647 records, and the full grounded expansion brings this to 78,880 before the final temporal addition. Counts refer to question records, so a video may contribute several examples.

\begin{table}[H]
\centering
\caption{RECAST-100k training records by task.}
\label{tab:app:dataset-composition}
\small
\setlength{\tabcolsep}{4pt}
\renewcommand{\arraystretch}{1.08}
\begin{adjustbox}{max width=\linewidth}
\begin{tabular}{@{}lrrr@{}}
\toprule
\textbf{Task} & \textbf{Original} & \textbf{Added} & \textbf{Final} \\
\midrule
Image QA & 5,000 & 0 & 5,000 \\
Temporal QA & 2,280 & 21,120 & 23,400 \\
Temporal QA (MCQ) & 2,904 & 0 & 2,904 \\
Spatiotemporal QA & 12,047 & 30,233 & 42,280 \\
General video QA & 15,000 & 11,416 & 26,416 \\
\midrule
\textbf{Total} & \textbf{37,231} & \textbf{62,769} & \textbf{100,000} \\
\bottomrule
\end{tabular}
\end{adjustbox}
\end{table}

\subsection{General Video QA}
\label{app:dataset-generalqa}
LLaVA-Video-178K~\citep{zhang2024videoinstruction}, Molmo2-HumanQA, and Molmo2-LongCapQA~\citep{clark2026molmo2} contribute 5,726, 3,784, and 1,906 training examples, respectively. Across training and held-out splits, we target 6,000, 4,000, and 2,000 records from these sources, totaling 8,000 four-option questions and 4,000 short answers. The teacher uses each source question as a topic seed and views up to 16 frames sampled at 2 FPS.

The same teacher answers each candidate twice without seeing the generated answer. For multiple-choice QA, both passes independently shuffle the options and must recover the intended choice with visual support. For short answers, a final answer-aware pass checks semantic agreement among the generated answer and both blind answers, as well as visual support. Failed support checks reject the candidate. Retries address only format or truncation errors.

We check for overlap using paths, recovered video identities, parent YouTube IDs, and full hashes when available. We export at most one question per video across both formats. A deterministic hash of each video path assigns 11,416 records to training and 584 to the held-out set.

\subsection{TimeLens Expansion}
\label{app:dataset-events}
We reuse event queries and local video intervals from TimeLens-100K~\citep{zhang2026timelens}, whose Gemini-2.5-Pro pipeline annotates events from CosMo-Cap, DiDeMo, HiREST, InternVid-VTime, and QuerYD. Among 19,466 video-level records, 52 have empty event lists. The others yield 96,586 valid events. We group videos using source IDs, aliases, and hashes. Excluding groups that overlap with protected training references, held-out sets, or evaluation data removes 3,542 events and leaves 93,044. The group-level split assigns 88,450 to training and 4,594 to the held-out set.

Temporal, object, action, and semantic cues route the training events into disjoint queues of 23,003 grounding-priority events, 57,069 temporal events, and 8,378 events for semantic review. The grounded training pool combines all 23,003 grounding-priority events with 26,997 temporal events, covering 17,742 parent-video identity groups. Another 1,176 grounded candidates retain their held-out assignment.

\paragraph{Grounded annotation.}\label{app:dataset-grounding}
The local teacher receives the source query, its interval, and up to 16 distinct frames. The default budget is eight event frames, four overview frames, and four neighboring frames within two seconds of the event. Sampling uses decoded indices and presentation timestamps, with repeated indices merged.

The first pass writes a question and answer, selects one supporting frame within the interval, and names one or two visible objects. The second localizes those objects on the clean frame with tight, axis-aligned boxes on a 0--1000 scale. We check numeric values, coordinate order, image bounds, and minimum size before normalizing to $[0,1]$. The third checks answer support, question clarity, event relevance, object relevance, and box correspondence. Acceptance requires all checks to pass and the selected image alone to support the answer. The same teacher runs the three passes with separate prompts and retries only for format or truncation errors.

Across 50,000 training and 1,176 held-out candidates, the run returns 30,900 accepted annotations, 15,796 semantic rejections, and 4,480 technical errors. Duplicate and conflict checks leave 30,893 exports, comprising 30,233 training records (60.47\% training export yield) and 660 held-out records.

\subsection{Data Export and Usage}
\label{app:dataset-temporal}
Of the 30,072 remaining temporal training events, 29,611 pass screening and 461 are held for review. Screening requires a nonempty query, a valid training assignment, and finite ordered boundaries. For accessible media, we also verify file identity, media header, duration, and alignment with the local timeline. We select 21,120 events using proportional source quotas and round-robin sampling over video identity groups within each source. The remaining 8,491 eligible events form a reserve.

Native temporal examples retain the source query, use the interval boundaries as the answer, and carry the \texttt{temporal QA} task type. Grounded QA adds the selected frame index, its timestamp, and its clean image path, with normalized boxes stored in \texttt{key\_items}. Native temporal and general video QA leave the key-frame and object fields empty. Provenance records retain source IDs, media paths, source intervals where applicable, and generation checks. Train and held-out assignments are separated by recovered video identity. Identity and hash checks exclude known overlap with protected evaluation and held-out videos. Visual near-duplicates may remain.

\paragraph{Training interface.}\label{app:dataset-training-interface}
Each record provides a visual input and question $x_i$ and annotations $a_i$. Student generation sees only $x_i$. Task labels select the answer, temporal, spatial, and format checks in $R(x_i,y_i,a_i)$. Available intervals, key frames, and boxes define the evidence views $E_i^m$ that the self-teacher uses to score tokens. The main-paper experiments use the full RECAST-100k export with the original task schema.

\section{Evaluation Details}
\label{app:evaluation-details}

We evaluate RECAST on video question answering, temporal grounding, and general video understanding. This section describes the benchmark groups, comparison models, and metrics underlying the main-paper results.

\subsection{Benchmarks}
\label{app:benchmarks}

\paragraph{Video reasoning and grounding.}
WorldSense~\citep{hong2026worldsense}, VideoMMMU~\citep{hu2026videommmu}, and LongVideo-Reason-eval (LRR)~\citep{chen2025longvila-r1} evaluate answer correctness on video questions. The main paper reports WorldSense Overall and Understanding, VideoMMMU Overall and Perception, and LRR accuracy. TVGBench~\citep{wang2025timer1} and Charades-STA~\citep{gao2017tall} evaluate temporal localization by comparing a predicted video interval with the reference interval. We report mIoU for both benchmarks and additionally report recall at three IoU thresholds for Charades-STA.

\paragraph{General video understanding.}
We report results on MVBench~\citep{li2024mvbench}, VideoMME~\citep{fu2025videomme} without subtitles, TempCompass~\citep{liu2024tempcompass} multiple-choice questions, and SEED-Bench~\citep{li2023seed}. These four benchmarks provide the general-understanding comparison in the main paper.


\subsection{Baselines}
\label{app:baselines}

The video reasoning and grounding comparison includes Qwen2.5-VL-7B~\citep{bai2025qwen}, VideoR1-7B~\citep{feng2025videor1}, VideoRFT-7B~\citep{wang2025videorft}, Open-o3-Video-7B~\citep{meng2026openo3}, VisionCoach-7B~\citep{lee2026visioncoach}, and VISD-7B~\citep{lin2026visd}. GPT-4o is included where a corresponding result is available. The Charades-STA comparison additionally covers general video models, temporal-grounding models, and grounded-reasoning models, as grouped in the main-paper table. \par
The general video understanding comparison includes LLaVA-OneVision-7B~\citep{li2024llavaonevision}, Qwen2.5-VL-7B, Video-R1, VideoChat-R1~\citep{li2025videochatr1}, VideoRFT, TW-GRPO~\citep{dang2025twgrpo}, and Open-o3-Video. A dash in the comparison tables denotes an unreported result; an average is not shown when one of its required component scores is missing.


\subsection{Evaluation Metrics}
\label{app:metrics}

For question answering, accuracy is the percentage of evaluated questions answered correctly. For temporal grounding, let $I_p$ and $I_g$ be the predicted and reference intervals. Their temporal intersection over union is
\begin{equation}
\operatorname{tIoU}(I_p,I_g)
=\frac{|I_p\cap I_g|}{|I_p\cup I_g|}.
\label{eq:app:tiou}
\end{equation}
Mean IoU averages this value over the evaluated queries. Charades-STA recall at threshold $\tau\in\{0.3,0.5,0.7\}$ is the percentage of queries for which $\operatorname{tIoU}(I_p,I_g)\geq\tau$.

The main reasoning table averages WorldSense Overall accuracy, VideoMMMU Overall accuracy, LRR accuracy, and TVGBench mIoU. The training-round curve in Fig.~1 of the main paper averages these four metrics and Charades-STA mIoU. The general video understanding table averages MVBench, VideoMME, TempCompass, and SEED-Bench when all four scores are available. Each mean weights its metrics equally.

\section{Limitations and Future Work}
\label{app:limitations}
RECAST uses task-specific evidence to guide its self-teacher during training. In RECAST-100k, event intervals specify when an event occurs, and grounded QA links each answer to a single evidence frame and its object boxes. This compact spatial view makes the annotated evidence easy to inspect for questions that can be answered from a decisive frame. Extending the spatial view to short frame sequences or object tracks could support evidence-conditioned training for questions that depend on changes across frames.

\section{Case Study}
\label{app:cases}

\begin{figure}[!htbp]
  \centering
  \includegraphics[trim=25bp 34bp 32bp 21bp,clip,width=\linewidth]{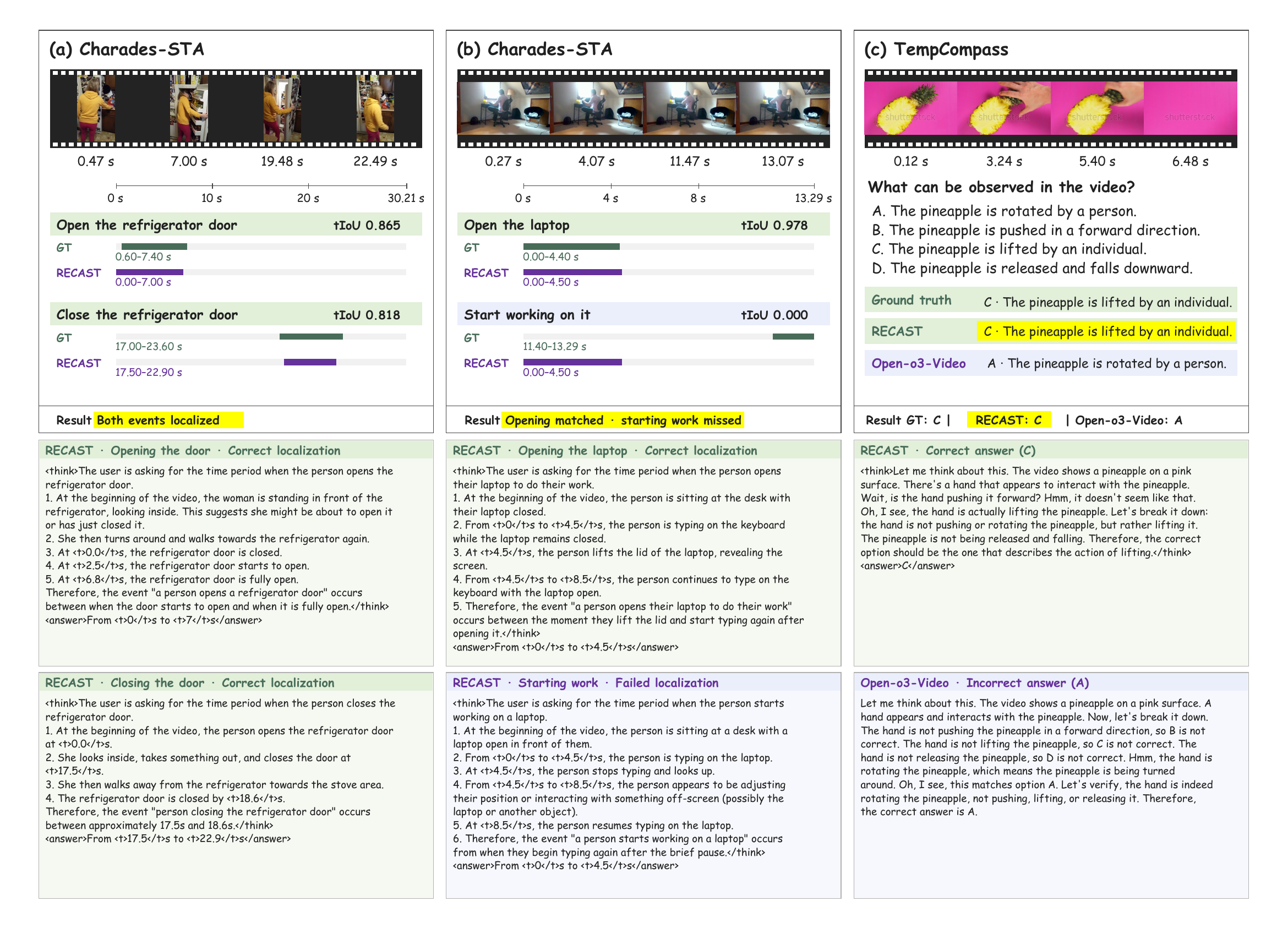}
  \caption{\textbf{Qualitative examples of temporal grounding and video QA.}
  (a) RECAST locates opening and closing the refrigerator door.
  (b) It predicts the same interval for opening a laptop and starting work.
  (c) RECAST selects lifting the pineapple, and Open-o3-Video selects rotation.
  The film strips show representative frames. Temporal IoU (tIoU) compares the final predicted interval with the ground truth.}
  \label{fig:app:recast-cases}
\end{figure}

\begin{figure}[H]
  \centering
  \includegraphics[trim=25bp 12bp 32bp 14bp,clip,width=\linewidth,height=.64\textheight,keepaspectratio]{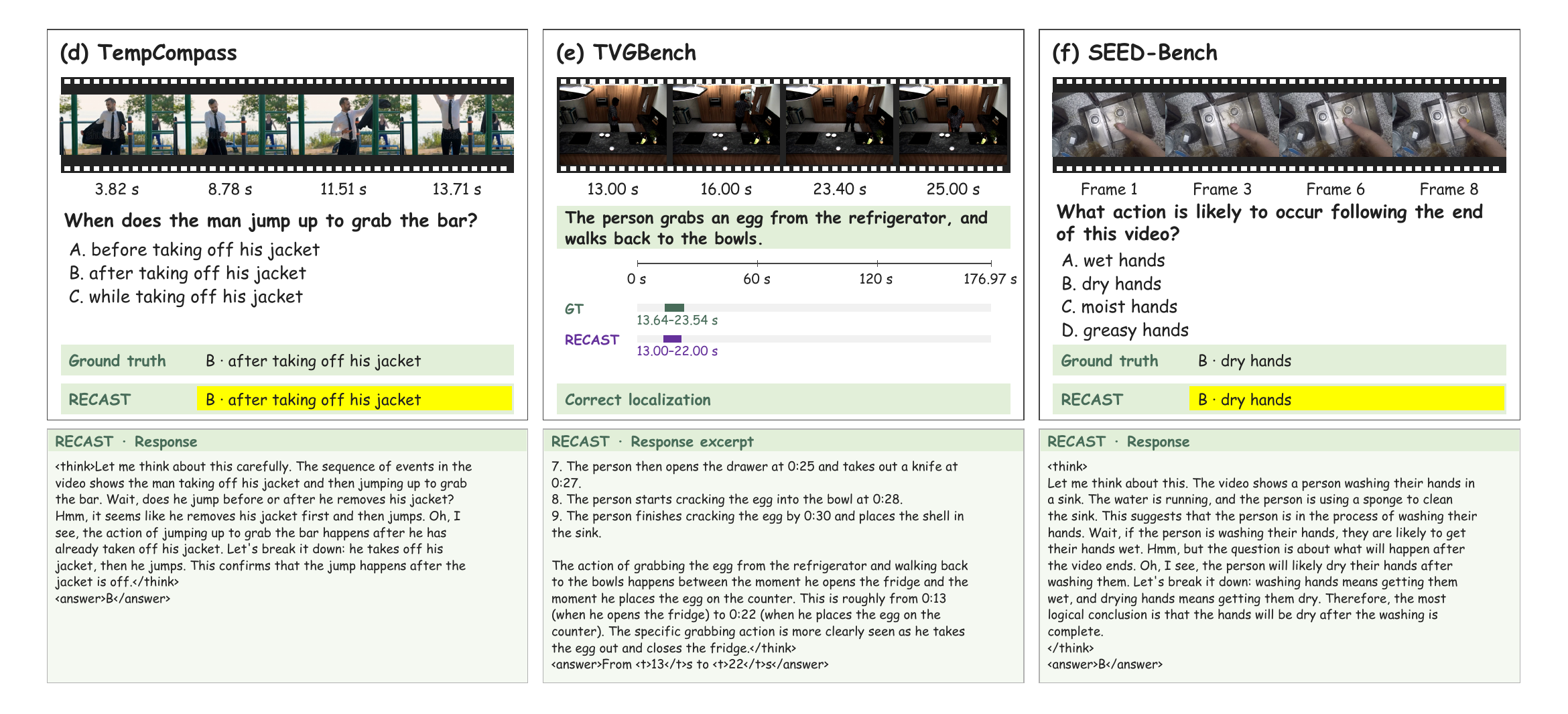}
  \caption{\textbf{Video examples of event ordering, temporal grounding, and action anticipation.}
  (d) RECAST places the jump after jacket removal on TempCompass.
  (e) On TVGBench, the predicted interval for retrieving an egg and returning to the bowls overlaps the reference interval.
  (f) On SEED-Bench, RECAST predicts hand drying after the clip ends.
  The lower panels show model responses, including an excerpt in (e).}
  \label{fig:app:recast-cases-more}
\end{figure}

Figures~\ref{fig:app:recast-cases} and~\ref{fig:app:recast-cases-more} present six examples of temporal grounding and video question answering. The Charades-STA and TVGBench examples come from the step-1073 preview. The TempCompass and SEED-Bench examples use RECAST at step 564. For the comparison in (c), both models received the same 16 sampled frames and used the same prompt and decoding settings.

\noindent\textbf{Separating events involving the same object.}
In (a), RECAST assigns distinct intervals to opening and closing the refrigerator door, with tIoUs of 0.865 and 0.818, respectively.

\noindent\textbf{Reusing an earlier interval for a later event.}
In (b), RECAST returns $[0.00,4.50]$\,s for both queries. The interval closely matches the laptop opening (tIoU 0.978) and has no overlap with the annotated start of work at $[11.40,13.29]$\,s. The response mentions a later resumption of typing, yet the final answer points to the opening.

\noindent\textbf{Distinguishing lifting from rotation.}
In (c), the hand lifts the pineapple out of view. RECAST selects the annotated option C (lifting). Open-o3-Video selects A (rotation).

\clearpage
\medskip
\noindent\textbf{Ordering events.}
In (d), jacket removal precedes the jump shown around 13.71\,s. RECAST selects B, the reference answer.

\noindent\textbf{Grounding a multi-step event.}
In (e), the predicted interval $[13.00,22.00]$\,s overlaps the annotated interval $[13.64,23.54]$\,s (tIoU 0.793). The prediction starts 0.64\,s early and ends 1.54\,s early.

\noindent\textbf{Anticipating the next action.}
In (f), the displayed frames show activity at the sink. RECAST predicts drying the hands after the clip ends, matching option B in the annotation.

\section{Training Details and Hyperparameters}
\label{app:training-details}

\subsection{Training Formulation and Recursive Schedule}
\label{app:formulation}

The student samples $y_i=(y_{i,1},\ldots,y_{i,T_i})$ from the visual input and question $x_i$, with history $h_{i,t}=(x_i,y_{i,<t})$. Annotations $a_i$ determine $R_i=R(x_i,y_i,a_i)$ and supply the available self-teacher evidence. At each step, the pre-update student $\pi_{\mathrm{old}}$ samples a group of responses per input. For group $g(i)$, let $\overline R_{g(i)}$ and $s_{g(i)}$ be its reward mean and sample standard deviation. The response-level advantage is
\begin{equation}
 A_i=\operatorname{sg}\left(
 \frac{R_i-\overline R_{g(i)}}{s_{g(i)}+\varepsilon_A}
 \right),
 \label{eq:app:adv}
\end{equation}
Here $\operatorname{sg}$ stops gradients, and $\pi_{\mathrm{ref}}$ is the frozen initial SFT policy.

As in Sec.~4 of the main paper, round $r$ alternates reward learning and teacher-guided learning. The pre- and post-reward snapshots define the self-teacher for the guided stage. Both guidance signals update the same student, whose final parameters start the next round.
\begin{equation}
 \theta_r^- \xrightarrow{\mathcal L_{\mathrm{task}}}
 \theta_r^+ \xrightarrow{\mathcal L_{\mathrm{guided}}}
 \theta_{r+1}^- .
 \label{eq:app:schedule}
\end{equation}

During guided step $s$, the current student $\theta_{r,s}$ supplies the rollout and pre-update continuation policy. The next step refreshes the sampled responses and local feedback, while the two reward-stage snapshots stay frozen until the next round.

\subsection{Policy-Extrapolated Self-Teacher}
\label{app:teacher}

\subsubsection{Policy Displacement under Shared Evidence}
The pre- and post-reward snapshots score the same history $h_{i,t}$ and evidence $E_i^m$. The available views $m\in\mathcal M_i$ are answer, answer--temporal, and answer--spatial. For each vocabulary token $v$,
\begin{equation}
 p_{r,i,t}^{m,\pm}(v)
 =\pi_{\theta_r^\pm}(v\mid h_{i,t},E_i^m),
 \qquad v\in\mathcal V.
 \label{eq:app:snapshots}
\end{equation}
We scale and clip the log-probability displacement, then normalize the extrapolated scores over the vocabulary $\mathcal V$.
\begin{align}
 \Delta_{r,i,t}^{m}(v)
 &=\operatorname{clip}\!\left(
 \alpha\log\frac{p_{r,i,t}^{m,+}(v)}{p_{r,i,t}^{m,-}(v)},
 -c_\Delta,c_\Delta\right),\label{eq:app:delta-teacher}\\
 q_{r,i,t}^{m}(v)
 &=\operatorname{softmax}_{v}\!\left(
 \log p_{r,i,t}^{m,+}(v)+\Delta_{r,i,t}^{m}(v)\right).
 \label{eq:app:teacher-view}
\end{align}
The scale $\alpha$ sets the extrapolation strength, and $c_\Delta$ bounds the change in each log weight. When $\alpha=0$, $q_{r,i,t}^m$ recovers the post-reward distribution for view $m$.

\subsubsection{Evidence Fusion and Snapshot Fallbacks}
The self-teacher averages the separately normalized distributions over valid views.
\begin{equation}
 q_{r,i,t}^{\mathrm{stage}}(v)
 =\frac{1}{|\mathcal M_i|}
 \sum_{m\in\mathcal M_i}q_{r,i,t}^{m}(v).
 \label{eq:app:fusion}
\end{equation}
We recompute $q_{r,i,t}^{\mathrm{stage}}$ on each new prefix. Failed stage-quality checks for format validity, length, or truncation disable extrapolation and select the pre-reward snapshot. If these checks pass but no valid nonzero-advantage update is available, we use the evidence-conditioned post-reward snapshot with extrapolation disabled.

\subsection{Structured Teacher-Guided Optimization}
\label{app:guidance}

Future-Aware Semantic Credit reweights advantages on sampled tokens. Outcome-Grounded Decision Refinement scores unselected alternatives at one position. Their outputs enter the joint loss in App.~\ref{app:optimization}.

\subsubsection{Future-Aware Semantic Credit}
\label{app:credit}

\paragraph{Event-aware chunk assignment.}
We partition valid response tokens into consecutive, nonoverlapping chunks $C_{i,1},\ldots,C_{i,K_i}$. Related entity, action, time, and location fields stay together where possible. Other text follows sentence or clause boundaries. The final answer forms a separate chunk. Invalid structural alignment falls back to deterministic length-based chunks. Let $k(t)$ denote the chunk of token $t$.

\paragraph{From sampled-token evidence to future-aware credit.}
We score each sampled token under the self-teacher relative to the pre-update student.
\begin{equation}
 \delta_{i,t}=\operatorname{sg}\!\left[
 \log q_{r,i,t}^{\mathrm{stage}}(y_{i,t})
 -\log\pi_{\mathrm{old}}(y_{i,t}\mid h_{i,t})
 \right].
 \label{eq:app:token-credit}
\end{equation}
We average within each chunk and combine the current and up to $H$ later chunks in the same completed response.
\begin{align}
 D_{i,k}&=\frac{1}{|C_{i,k}|}
 \sum_{t\in C_{i,k}}\delta_{i,t},
 \label{eq:app:chunk-credit}\\
 F_{i,k}&=\frac{
 \sum_{j=k}^{\min(K_i,k+H)}\gamma^{j-k}D_{i,j}}
 {\sum_{j=k}^{\min(K_i,k+H)}\gamma^{j-k}}.
 \label{eq:app:future}
\end{align}
The denominator normalizes the $\gamma$-discounted window at the end of the response. We center the credit by its token-weighted response mean to reweight $A_i$.
\begin{align}
 \overline F_i&=\frac{\sum_{k=1}^{K_i}|C_{i,k}|F_{i,k}}
 {\sum_{k=1}^{K_i}|C_{i,k}|},
 \label{eq:app:credit-center}\\
 w_{i,t}&=\operatorname{clip}\!\left(
 \exp\!\left[\operatorname{sign}(A_i)
 (F_{i,k(t)}-\overline F_i)\right],
 1-c_w,1+c_w\right),\label{eq:app:credit-weight}\\
 \widetilde A_{i,t}&=\operatorname{sg}(A_iw_{i,t}).
 \label{eq:app:adjusted-adv}
\end{align}
Unusable evidence gives $w_{i,t}=1$. The positive weight preserves the sign of the reward-derived advantage.

\subsubsection{Outcome-Grounded Decision Refinement}
\label{app:branch}

\paragraph{One disagreement-selected position.}
Let $\mathcal C_{i,t}=\{c_1,\ldots,c_k\}$ be the top-$k$ tokens of $p_{i,t}^{\mathrm{old}}=\pi_{\mathrm{old}}(\cdot\mid h_{i,t})$. We project both policies onto this common set and one residual bucket.
\begin{equation}
 \mathcal B_{\mathcal C}(p)
 =\left(p(c_1),\ldots,p(c_k),
 1-\sum_{j=1}^{k}p(c_j)\right).
 \label{eq:app:bucket}
\end{equation}
We score disagreement by forward KL on the common buckets.
\begin{equation}
 u_{i,t}=D_{\mathrm{KL}}\!\left(
 \mathcal B_{\mathcal C_{i,t}}(q_{r,i,t}^{\mathrm{stage}})
 \,\middle\|\,
 \mathcal B_{\mathcal C_{i,t}}(p_{i,t}^{\mathrm{old}})
 \right).
 \label{eq:app:disagreement}
\end{equation}
The eligible set $\mathcal E_i$ requires valid response tokens and teacher scores, sufficient top-$k$ coverage and disagreement, and continuation budget. We then select one position per response.
\begin{equation}
 t_i^*=\underset{t\in\mathcal E_i}
 {\operatorname*{arg\,max}}\ u_{i,t}.
 \label{eq:app:position}
\end{equation}
An empty $\mathcal E_i$ skips the branch.

\paragraph{Outcome estimates from fresh student continuations.}
With the visual input and prefix $y_{i,<t_i^*}$ fixed, we force each tested candidate $c_j$ and sample fresh suffixes from the pre-update student's decoding distribution $\mu_{\mathrm{old}}$. The original task reward scores the completed responses.
\begin{align}
 z_{j,n}&\sim\mu_{\mathrm{old}}(\cdot\mid
 x_i,y_{i,<t_i^*},c_j),\nonumber\\
 \widetilde y_{j,n}&=y_{i,<t_i^*}\oplus c_j\oplus z_{j,n},
 \label{eq:app:continuation}\\
 Q_j&=\frac{1}{N_j}\sum_{n=1}^{N_j}
 R(x_i,\widetilde y_{j,n},a_i).
 \label{eq:app:branch-value}
\end{align}
The average uses only fresh, successfully scored continuations. A terminal candidate is scored directly.

\paragraph{Reward-grounded local probability target.}
We anchor the target to $p(v)=\pi_{\mathrm{old}}(v\mid h_{i,t_i^*})$. Let $\mathcal J$ be the scored candidates. Their coverage and probability-weighted baseline are
\begin{equation}
 \omega=\sum_{j\in\mathcal J}p(c_j),\qquad
 Q_0=\frac{\sum_{j\in\mathcal J}p(c_j)Q_j}{\omega}.
 \label{eq:app:baseline}
\end{equation}
With at least two scored candidates and $\omega>0$, we apply the bounded bias
\begin{equation}
 b(v)=\begin{cases}
 \kappa\omega\,\operatorname{clip}(Q_j-Q_0,-c_Q,c_Q),
 &v=c_j,\ j\in\mathcal J,\\
 0,&\text{otherwise}.
 \end{cases}
 \label{eq:app:bias}
\end{equation}
The full-vocabulary branch target is
\begin{align}
 q_i^{\mathrm{branch}}(v)
 &=\frac{p(v)e^{b(v)}}{Z_i},
 \label{eq:app:target}\\
 Z_i&=\sum_{u\in\mathcal V}p(u)e^{b(u)}
 =1+\sum_{j\in\mathcal J}p(c_j)(e^{b(c_j)}-1).
 \label{eq:app:normalizer}
\end{align}
Unevaluated tokens retain their relative probabilities, and $\omega$ scales the tilt by candidate coverage. Project both the target and the current student at $t_i^*$ onto $\mathcal C_{i,t_i^*}$ plus its residual bucket to obtain $\widehat q_i^{\mathrm{branch}}$ and $\widehat\pi_{\theta,i}$. For responses $\mathcal I_B$ with a valid branch,
\begin{equation}
 \mathcal L_{\mathrm{branch}}
 =\frac{1}{|\mathcal I_B|}\sum_{i\in\mathcal I_B}
 D_{\mathrm{KL}}\!\left(
 \widehat q_i^{\mathrm{branch}}\,\middle\|\,
 \widehat\pi_{\theta,i}\right).
 \label{eq:app:branch-loss}
\end{equation}
We omit $\mathcal L_{\mathrm{branch}}$ if $\mathcal I_B=\varnothing$. The teacher selects the position, and the task rewards determine the candidate preferences.

\subsection{Joint Objectives and Implementation}
\label{app:optimization}

Both stages use a group-relative clipped policy loss and fixed reference regularization. Teacher-guided learning reweights the advantages and adds the local branch loss.

\subsubsection{Full Clipped Policy Objective}
For a sampled token, define its probability ratio and clipped group-relative surrogate:
\begin{align}
 \rho_{i,t}(\theta)
 &=\frac{\pi_\theta(y_{i,t}\mid h_{i,t})}
 {\pi_{\mathrm{old}}(y_{i,t}\mid h_{i,t})},
 \label{eq:app:ratio}\\
 \Psi(\rho,B)&=\min\{\rho B,
 \operatorname{clip}(\rho,1-\epsilon,1+\epsilon)B\}.
 \label{eq:app:ppo-surrogate}
\end{align}
Following GRPO~\citep{shao2024deepseekmath}, we average valid tokens within each response and then average over valid responses $\mathcal I$.
\begin{equation}
 \mathcal L_{\mathrm{pg}}(B)
 =-\frac{1}{|\mathcal I|}\sum_{i\in\mathcal I}
 \frac{1}{T_i}\sum_{t=1}^{T_i}
 \Psi(\rho_{i,t}(\theta),B_{i,t}).
 \label{eq:app:pg}
\end{equation}
Reward learning uses $B_{i,t}=A_i$. Teacher-guided learning uses $B_{i,t}=\widetilde A_{i,t}$.

\subsubsection{Fixed Reference Regularization}
At sampled tokens, we clip the log ratio $\ell_{i,t}=\operatorname{clip}(\log\pi_{\mathrm{ref}}-\log\pi_\theta,-20,20)$, with both probabilities evaluated at $y_{i,t}$ under $h_{i,t}$. The forward term is
\begin{equation}
 K_{i,t}^{(3)}=\operatorname{clip}\!\left(
 e^{\ell_{i,t}}-\ell_{i,t}-1,0,10\right),
 \label{eq:app:ref-k3}
\end{equation}
We average over valid response tokens in the policy-loss groups for $\mathcal L_{\mathrm{ref}}$. Backpropagation uses the $k_2$ surrogate with detached coefficient $-\ell_{i,t}$.

\subsubsection{Stage Losses and Detached Quantities}
The full task-stage and guided-stage objectives are
\begin{align}
 \mathcal L_{\mathrm{task}}
 &=\mathcal L_{\mathrm{pg}}(A)
 +\lambda_{\mathrm{ref}}\mathcal L_{\mathrm{ref}},
 \label{eq:app:task-loss}\\
 \mathcal L_{\mathrm{guided}}
 &=\mathcal L_{\mathrm{pg}}(\widetilde A)
 +\lambda_{\mathrm{ref}}\mathcal L_{\mathrm{ref}}
 +\lambda_B\mathcal L_{\mathrm{branch}}.
 \label{eq:app:guided-loss}
\end{align}
The old sampling policy, both snapshots, rewards, chunk assignments, credit weights, and branch targets are detached. Each eligible optimizer minibatch combines the guided loss terms and updates only the current student.

\subsubsection{Training Settings and Validity Rules}
Table~\ref{tab:app:training-hyperparameters} gives the full configuration. Snapshot fallback follows App.~\ref{app:teacher}, and branch eligibility and skipping follow App.~\ref{app:branch}.

\raggedbottom
\subsection{Mathematical Properties}
\label{app:proofs}

\subsubsection{The Branch Target Solves a KL-Regularized Problem}
\noindent\textbf{Proposition 1.}
Let $p$ be a vocabulary distribution with $p(v)>0$, let $b(v)$ be finite, and set $Z=\sum_v p(v)e^{b(v)}$. Then $q^*(v)=p(v)e^{b(v)}/Z$ uniquely maximizes
\begin{equation}
 J(q)=\sum_v q(v)b(v)-D_{\mathrm{KL}}(q\|p),
 \qquad q\in\Delta(\mathcal V).
 \label{eq:app:variational}
\end{equation}
\noindent\textit{Proof.} Substituting $\log q^*(v)=\log p(v)+b(v)-\log Z$ gives
\begin{equation}
 J(q)=\log Z-D_{\mathrm{KL}}(q\|q^*).
 \label{eq:app:kl-proof}
\end{equation}
KL is nonnegative and zero only at $q=q^*$.\hfill$\square$

\noindent\textbf{Corollary 1.}
For tokens $u,v$ with positive student probability, the normalizer cancels:
\begin{equation}
 \log\frac{q^*(u)}{q^*(v)}
 =\log\frac{p(u)}{p(v)}+b(u)-b(v).
 \label{eq:app:odds}
\end{equation}

\subsubsection{Advantage Sign and Weight Bounds}
\noindent\textbf{Proposition 2.} If $0<c_w<1$, then
\begin{equation}
 1-c_w\le w_{i,t}\le1+c_w,
 \label{eq:app:boundw}
\end{equation}
and for $A_i\ne0$,
\begin{equation}
 \begin{aligned}
  \operatorname{sign}(\widetilde A_{i,t})&=\operatorname{sign}(A_i),\\
  (1-c_w)|A_i|&\le|\widetilde A_{i,t}|\le(1+c_w)|A_i|.
 \end{aligned}
 \label{eq:app:adv-bounds}
\end{equation}
\noindent\textit{Proof.} Clipping bounds $w_{i,t}$ and makes it positive. Thus $\widetilde A_{i,t}=w_{i,t}A_i$ preserves the sign, scales $|A_i|$ within the stated factors, and equals zero when $A_i=0$. The claim concerns the policy-gradient advantage.\hfill$\square$

\subsubsection{Aggregation Properties of Future Credit}
\noindent\textbf{Proposition 3.}
For $\gamma\ge0$ and a nonempty credit window $W_{i,k}=\{k,\ldots,\min(K_i,k+H)\}$ with a positive normalizer, $F_{i,k}$ is a convex combination of the window credits. Hence
\begin{equation}
 \min_{j\in W_{i,k}}D_{i,j}
 \le F_{i,k}\le
 \max_{j\in W_{i,k}}D_{i,j}.
 \label{eq:app:future-bounds}
\end{equation}
\noindent\textit{Proof.} The normalized weights proportional to $\gamma^{j-k}$ are nonnegative and sum to one, with $\gamma^0=1$ also at $\gamma=0$.\hfill$\square$

\subsubsection{Information Kept by Candidate Buckets}
\noindent\textbf{Proposition 4.}
For distributions $q,p$ with $q$ absolutely continuous with respect to $p$, a common candidate set and a single residual bucket give
\begin{equation}
 D_{\mathrm{KL}}\!\left(
 \mathcal B_{\mathcal C}(q)\middle\|\mathcal B_{\mathcal C}(p)
 \right)\le D_{\mathrm{KL}}(q\|p).
 \label{eq:app:bucket-bound}
\end{equation}
\noindent\textit{Proof.}
On $S=\mathcal V\setminus\mathcal C$, the log-sum inequality gives
\begin{equation}
 \sum_{v\in S}q(v)\log\frac{q(v)}{p(v)}
 \ge q(S)\log\frac{q(S)}{p(S)},
 \label{eq:app:logsum}
\end{equation}
where the right-hand side is zero when $q(S)=0$. The candidate terms are unchanged, so adding them gives the bound.\hfill$\square$

\par\medskip
\begingroup
\newcommand{\recastparamgroup}[1]{%
  \multicolumn{2}{@{}l@{}}{%
    \setlength{\fboxsep}{2pt}%
    \colorbox[RGB]{232,237,247}{%
      \makebox[\dimexpr\columnwidth-2\fboxsep\relax][l]{\strut\emph{#1}}}}\\}
\noindent\begin{minipage}{\columnwidth}
\subsection{Hyperparameter Summary}
\label{app:hyperparameters}
Table~\ref{tab:app:training-hyperparameters} gives the main-paper training configuration.

\centering
\captionsetup{type=table}
\caption{Training hyperparameters for RECAST-7B.}
\label{tab:app:training-hyperparameters}
\small
\setlength{\tabcolsep}{4pt}
\renewcommand{\arraystretch}{0.91}
\begin{tabular*}{\columnwidth}{@{\extracolsep{\fill}}lr@{}}
\toprule
\textbf{Setting} & \textbf{Value} \\
\midrule
\recastparamgroup{Model and input}
Initialization & Open-o3-Video-SFT-7B \\
Training data & RECAST-100k (100,000) \\
Base video sampling & 2 FPS target, max. 16 frames \\
Student prompt / response & 16,384 / 1,024 tokens \\
Teacher prompt limit & 18,432 tokens \\
\midrule
\recastparamgroup{Optimization and rollout}
Hardware / training duration & 32 A800 GPUs / 1 epoch \\
Rollout steps & 1,563 \\
Prompts per rollout & 64 \\
Prompt groups per update & 16 \\
Responses per prompt & 4 \\
Task / guided steps per round & 12 / 12 \\
Optimizer / initial LR & AdamW / $10^{-6}$ \\
Betas / weight decay & $(0.9,0.999)$ / 0.01 \\
Learning-rate schedule & Cosine, no warmup \\
Gradient-norm cap & 5.0 \\
Policy-ratio clipping & $[0.8,1.2]$ \\
Rollout $T$ / top-$p$ / top-$k$ & 1.0 / 0.95 / 50 \\
Rollout engine & vLLM \\
Precision / sharding & BF16 / FSDP \\
Shuffle seed & 1 \\
Checkpoint interval / retention & 12 steps / 25 \\
\bottomrule
\end{tabular*}
\end{minipage}\par\medskip

\noindent\begin{minipage}{\columnwidth}
\centering
\small
Table~\ref{tab:app:training-hyperparameters} (continued).\par\smallskip
\setlength{\tabcolsep}{4pt}
\renewcommand{\arraystretch}{0.91}
\begin{tabular*}{\columnwidth}{@{\extracolsep{\fill}}lr@{}}
\toprule
\textbf{Setting} & \textbf{Value} \\
\midrule
\recastparamgroup{RECAST-specific parameters}
GRPO stabilizer $\varepsilon_A$ & $10^{-4}$ \\
Teacher $\alpha$ / $c_\Delta$ & 0.5 / 1 \\
Teacher temperature & 1 \\
FASC $H$ / $\gamma$ / $c_w$ & 2 / 0.8 / 0.2 \\
ODR candidates per position & 4 \\
ODR continuations per candidate & 2 (nonterminal) \\
ODR coverage / KL threshold & 0.5 / 0.05 nats \\
ODR $\kappa$ / $c_Q$ & 1 / 1 \\
Reference-KL weight $\lambda_{\mathrm{ref}}$ & 0.04 \\
ODR loss weight $\lambda_B$ & 1 \\
\bottomrule
\end{tabular*}
\end{minipage}\par
\endgroup
\end{document}